\PassOptionsToPackage{round,authoryear}{natbib}
\documentclass{article}

\usepackage{arxiv}
\usepackage[utf8]{inputenc}
\usepackage[T1]{fontenc}
\usepackage{amsmath,amssymb}
\usepackage{graphicx}
\usepackage{booktabs,longtable,array,multirow}
\usepackage{microtype}
\usepackage{natbib}
\usepackage{url,xurl}
\usepackage{hyperref}
\usepackage{placeins}
\usepackage{float}
\usepackage{caption}
\hypersetup{hidelinks}
\graphicspath{{./}}
\makeatletter
\@addtoreset{figure}{section}
\@addtoreset{table}{section}
\makeatother

\providecommand{\keywords}[1]{\par\noindent\textbf{Keywords:} #1\par}

\title{BrainNet Studio: A Unified Toolkit for Brain Network Construction, Intelligent Analysis, and Visualization}
\author{\parbox{0.95\textwidth}{\centering\normalfont
\textbf{Xiwei Zeng\textsuperscript{1}, Shengrong Li\textsuperscript{1}, Yiheng Liu\textsuperscript{1},}\\[2pt]
\textbf{Chunwei Tian\textsuperscript{2}, Daoqiang Zhang\textsuperscript{1}, Qi Zhu\textsuperscript{1,*}}\\[6pt]
{\small
\textsuperscript{1}College of Artificial Intelligence,\\
Nanjing University of Aeronautics and Astronautics, Nanjing, China\\[3pt]
\textsuperscript{2}School of Computer Science and Technology,\\
Harbin Institute of Technology, Harbin, China\\[6pt]}
{\footnotesize
\href{mailto:xwzeng@nuaa.edu.cn}{\texttt{xwzeng@nuaa.edu.cn}},
\href{mailto:lisrong@nuaa.edu.cn}{\texttt{lisrong@nuaa.edu.cn}},
\href{mailto:sx2524003@nuaa.edu.cn}{\texttt{sx2524003@nuaa.edu.cn}},\\
\href{mailto:chunweitian@hit.edu.cn}{\texttt{chunweitian@hit.edu.cn}},
\href{mailto:dqzhang@nuaa.edu.cn}{\texttt{dqzhang@nuaa.edu.cn}},
\href{mailto:zhuqi@nuaa.edu.cn}{\texttt{zhuqi@nuaa.edu.cn}}\\[3pt]
\textsuperscript{*}Corresponding author.}
}}
\hypersetup{pdfauthor={Xiwei Zeng, Shengrong Li, Yiheng Liu, Chunwei Tian, Daoqiang Zhang, Qi Zhu}}

\begin{document}
\maketitle

\raggedbottom
\renewcommand{\topfraction}{0.92}
\renewcommand{\bottomfraction}{0.85}
\renewcommand{\textfraction}{0.06}
\renewcommand{\floatpagefraction}{0.75}
\setcounter{topnumber}{3}
\setcounter{bottomnumber}{2}
\setcounter{totalnumber}{4}
\setlength{\textfloatsep}{12pt plus 2pt minus 2pt}
\setlength{\intextsep}{10pt plus 2pt minus 2pt}
\setlength{\floatsep}{10pt plus 2pt minus 2pt}

\begin{abstract}
Brain networks characterize structural and functional relationships among brain regions and provide an important basis for investigating brain organization, cognitive processes, disease-related alterations, and brain-computer interfaces. Brain activity, however, exhibits time-varying topology and higher-order spatiotemporal dependencies that cannot be adequately represented by conventional static networks. Existing analytical tools primarily focus on static connectomes and provide limited support for integrating dynamic network modeling with modern graph and sequence learning methods. Moreover, the lack of unified designs for network construction, feature representation, and downstream analysis requires researchers to manually assemble complex multistage pipelines. To address these limitations, we developed BrainNet Studio, an integrated toolkit for static and dynamic brain network analysis in cognitive neuroscience and exploratory studies of brain disorders. BrainNet Studio provides a unified workflow encompassing network construction, feature extraction, predictive modeling, candidate biomarker identification, visualization, and assisted interpretation. It integrates dynamic brain network methods with deep learning, graph neural networks, and spatiotemporal sequence models to support multiple classification tasks and the identification of discriminative brain regions and connections. The toolkit also incorporates a large language model to generate researcher-verifiable summaries of functional connectivity, structural connectivity, and structure--function coupling results at both individual and group levels. By integrating 27 established and emerging algorithms within a consistent computational framework, it enables researchers to configure analytical tasks, execute and compare methods, inspect outputs, and extend existing functionality without repeatedly constructing application-specific pipelines. BrainNet Studio therefore provides a practical and extensible platform for connectome analysis, supporting research in cognitive neuroscience, brain disorders, and brain-computer interfaces. The toolkit is publicly available at https://github.com/xbrainnet/Brainnet-Studio.
\end{abstract}



\section{Introduction}

Brain function arises from complex structural and functional interactions among brain regions rather than the isolated activity of individual regions. These interactions form a multilevel, dynamically coupled neural network system. Brain connectivity analysis represents distinct brain regions as network nodes. Network edges encode anatomical fiber pathways and functional dependencies between regions. A standardized network analysis framework then quantifies the organizational patterns and interactions of the brain as a whole. Through this modeling framework, brain networks enable precise characterization of structural and functional relationships across multiple dimensions. They serve as important tools for uncovering the principles of brain organization and function, indicating the neural mechanisms of higher cognitive functions, and identifying pathological abnormalities in brain disorders. Brain network analysis provides a theoretical foundation and methodological framework for cognitive neuroscience and precision research on brain disorders. It is now widely applied to investigations of cognitive mechanisms, the pathophysiology of brain disorders, and brain--computer interfaces (Bassett and Sporns, 2017; Fotiadis et al., 2024).

Most existing brain network analysis tools rely on static network models for network construction and feature extraction. They estimate average statistical associations across the entire functional imaging time series to generate a single, time-invariant connectivity matrix. This matrix characterizes the steady-state network properties of the brain over the scanning period. However, human functional connectivity is inherently dynamic. Even at rest, functional coupling strengths between brain regions and the spatiotemporal patterns of integration and segregation across large-scale networks continuously evolve. Consequently, temporal averaging in static network representations obscures key dynamic information, including transient connectivity fluctuations and network state transitions. These representations therefore cannot faithfully capture the dynamic mechanisms underlying brain function. However, dynamic brain networks estimate connectivity within successive time windows, capturing the temporal evolution of interregional interactions and changes in network topology (Lurie et al., 2020). Beyond the simple lower-order associations represented by static networks, dynamic brain networks encompass more complex spatiotemporal coupling and higher-order topological dependencies. Characterizing their transient state changes, temporal dependencies, and cross-scale interactions relies heavily on sophisticated methods for higher-order spatiotemporal graph learning.

Rapid advances in cognitive neuroscience and neuroimaging have led to the development of specialized toolkits addressing diverse needs in brain connectome research. For network topology analysis, the Brain Connectivity Toolbox (BCT) (Rubinov and Sporns, 2010) provides a comprehensive collection of quantitative complex network measures, enabling accurate computation of diverse topological properties. For brain connectivity analysis, CONN (Whitfield-Gabrieli and Nieto-Castanon, 2012) provides a workflow for functional connectivity estimation and statistical analysis of resting-state and task-based fMRI data. PANDA (Cui et al., 2013) focuses on diffusion MRI processing to support structural brain network construction and analysis. For network statistics and between-group comparisons, GRETNA (Wang et al., 2015) integrates resting-state fMRI preprocessing, brain network construction, graph-theoretical metric computation, and group difference analysis into a comprehensive workflow. BRAPH (Mijalkov et al., 2017) combines network analysis, statistical testing, and visualization of results within a unified environment. GraphVar (Waller et al., 2018) supports machine learning based on brain connectivity features and model performance evaluation. BrainNetClass (Zhou et al., 2020) integrates brain network construction, feature selection, classification model training, and cross-validation into a complete workflow. NeuroPycon (Meunier et al., 2020) uses workflow management to automate multimodal brain connectivity analyses.

Although existing tools meet brain network analysis needs in specific settings, they retain several key technical limitations. Most mainstream tools rely on static network models for network construction and feature extraction. They characterize brain network organization using connectivity patterns averaged over the entire recording period. Dedicated capabilities for characterizing and visualizing the temporal evolution of dynamic networks remain limited. Furthermore, existing tools lack effective methods for extracting and analyzing higher-order spatiotemporal dependencies, complex topological interactions, and deep representations within dynamic brain networks. These limitations substantially hinder the application of advanced dynamic brain network algorithms to understanding cognitive mechanisms and achieving precise clinical diagnosis of brain disorders. An integrated analytical framework that combines dynamic modeling, higher-order spatiotemporal feature learning, and intelligent visualization is therefore needed. Such a framework would support the continued development and application of emerging brain network analysis methods.

To address these limitations, we designed a dedicated toolkit for dynamic brain network analysis. It integrates advanced algorithms for dynamic connectivity construction, higher-order spatiotemporal feature extraction, biomarker discovery, and dynamic brain network analysis. During network construction, visualization components display data distributions and the temporal evolution of brain networks. For feature analysis, the toolkit incorporates 27 brain network feature extraction methods. These methods characterize complex spatiotemporal network dynamics beyond the limitations of conventional static models and lower-order topological analyses. They support automated classification and the investigation of potential neural mechanisms. For interpretability analysis, model explanation algorithms map model decisions to key brain connections and regions. This enables visual exploration of candidate brain biomarkers and potential neural mechanisms.

In addition, recent advances in large language models (LLMs) have enhanced their capabilities in integrating complex information, linking domain-specific knowledge, and generating natural language (Singhal et al., 2023). These advances offer new approaches to the semantic organization and assisted interpretation of brain network analysis results. In this study, we integrate an LLM into the toolkit to synthesize findings, assist mechanistic interpretation, and generate supporting reports. The module uses structured, traceable analysis results as input and supports the interpretation of findings at individual or group levels (Gao et al., 2023). This extends the capabilities of conventional brain network toolkits to result interpretation and knowledge translation.

The resulting toolkit provides an integrated platform for dynamic brain network analysis in clinical and cognitive neuroscience research. Its principal innovations and contributions are as follows:

1. The toolkit establishes an extensible platform that seamlessly integrates the entire workflow of brain network construction, higher-order feature extraction, visualization, and analytical report generation. All core analytical procedures can be accomplished on this unified platform without switching between disparate tools or pipelines. These components provide systematic support for basic cognitive neuroscience, exploratory clinical research on brain disorders, and brain--computer interface investigations. 

2. The toolkit provides an interactive graphical user interface that encapsulates complex dynamic brain network algorithms and workflows. Users can conduct analyses without writing code or configuring complex runtime environments. This substantially lowers the technical barriers to advanced analytical methods and facilitates their use by clinicians and cognitive neuroscience researchers. The platform can improve data analysis efficiency and the reproducibility of experimental results, providing an accessible, specialized environment for dynamic brain network analysis.

3. The toolkit integrates 27 specialized algorithms for brain network feature analysis and supports both static and dynamic connectome analyses. These algorithms support dynamic connectivity state modeling, automated classification across multiple tasks, and the extraction of complex spatiotemporal topological features. The unified framework accommodates diverse analytical requirements.

4. The toolkit innovatively incorporates an LLM-based analysis module grounded in structured, traceable data. It synthesizes functional connectivity, structural connectivity, and structure--function coupling results at both individual and group levels. The module assists mechanistic interpretation and generates standardized reports, integrating data analysis with the interpretation of neuroscientific knowledge.

\noindent\begin{minipage}{\linewidth}
\section{Toolkit Functional Modules}

\centering
\includegraphics[width=\linewidth,height=0.72\textheight,keepaspectratio]{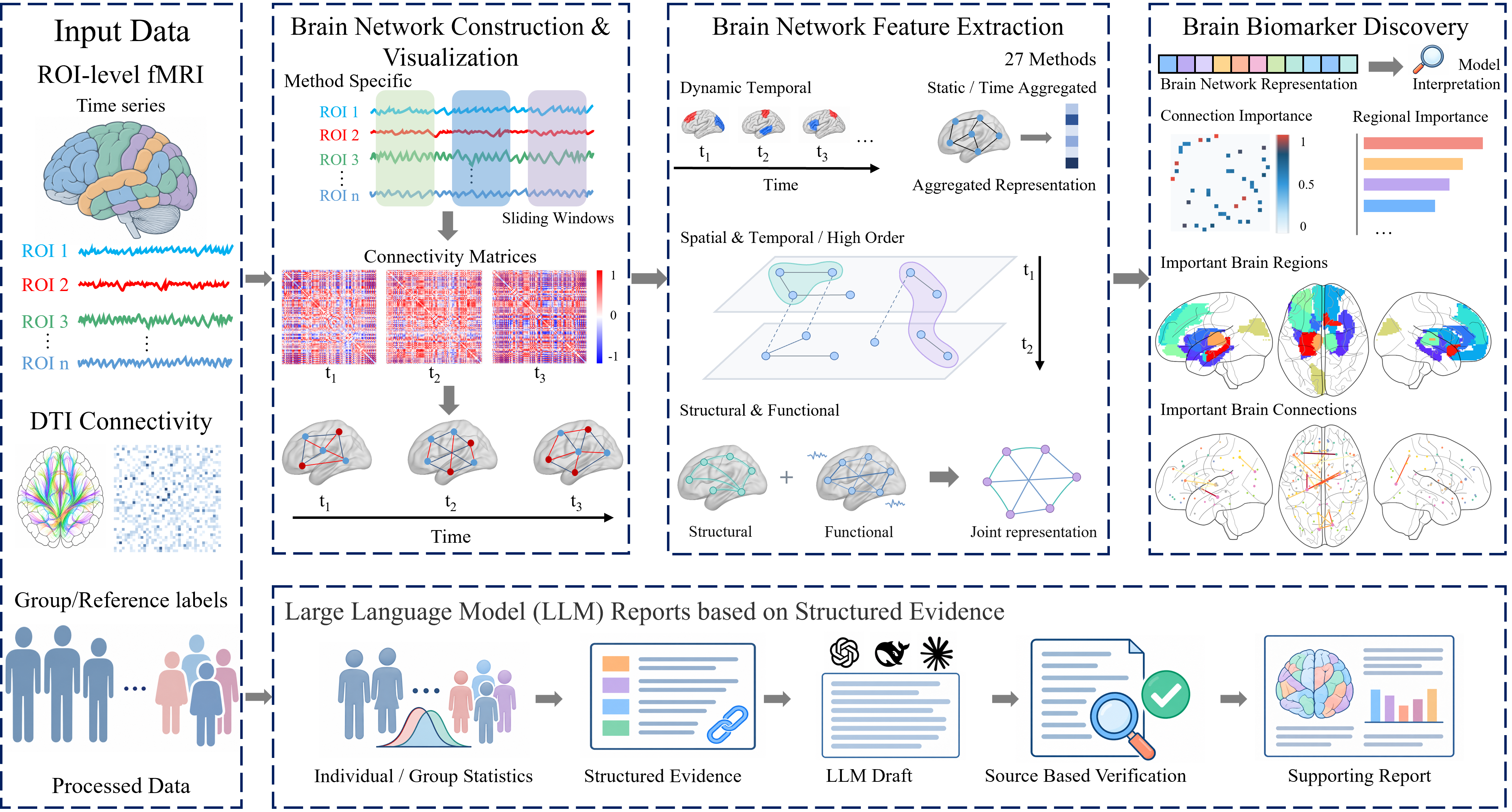}
\captionsetup{hypcap=false}
\captionof{figure}{Overview of the toolkit's functional modules and analysis workflows. The upper workflow connects brain network construction and visualization, feature extraction, and candidate brain biomarker discovery. The lower, relatively independent workflow organizes individual- or group-level results into structured evidence for LLM-assisted reporting and verification against source data. Arrows indicate processing flow. ROI, region of interest; LLM, large language model.}
\label{fig:2-1}
\end{minipage}
\par

The toolkit comprises four functional modules centered on the construction, representation, and interpretation of dynamic brain networks. These modules cover dynamic brain network construction and visualization, brain network feature extraction, brain biomarker discovery, and large language model (LLM) reports based on structured evidence. The construction and visualization module supports both static and dynamic brain networks. It generates network representations according to the construction strategy of the selected method and visualizes connectivity patterns, network topology, and the temporal evolution of dynamic networks. The feature extraction module integrates 27 analytical methods for extracting advanced brain network features, including complex spatiotemporal dependencies, higher-order topological relationships, and multimodal associations. It also includes conventional statistical learning and machine learning methods, providing broad methodological support for downstream tasks. The biomarker discovery module maps model decisions to specific brain regions and connections. It identifies and visualizes key regions and connections, providing a basis for candidate brain biomarker discovery and investigations of related neural mechanisms. These three modules form a sequential workflow from network modeling and feature learning to biological interpretation. The LLM reporting module operates relatively independently of this workflow and accepts processed unimodal or multimodal data. Depending on the input modalities and analysis mode, it computes brain network metrics at the individual or group level and organizes them into structured evidence. The LLM then synthesizes this evidence, provides assisted interpretation, and generates a brain network analysis report. Figure 2-1 provides an overview of the functional modules and their analysis workflows. The following sections describe each module.

\subsection{Brain Network Construction and Visualization}

Brain network construction converts regional brain signals into computational network representations. It provides the foundation for subsequent feature extraction, model training, and biomarker discovery. The primary input is region-of-interest (ROI)-level functional magnetic resonance imaging (fMRI) time series obtained after regional signal extraction. Brain regions are defined as network nodes, while interregional functional statistical dependencies or anatomical connections are defined as network edges (Bassett and Sporns, 2017). Different analytical methods require different temporal scales, connectivity definitions, and forms of network organization. The module therefore generates static networks, dynamic network sequences, or multilayer networks according to the construction strategy of the selected method.

The network construction process is described in greater detail below. For participant $s$, let the ROI-level fMRI time series be $\mathbf{X}_{s}\in\mathbb{R}^{N\times M}$ and the optional structural connectivity matrix be $\mathbf{S}_{s}\in\mathbb{R}^{N\times N}$. Here, $N$ and $M$ denote the numbers of brain regions and time points, respectively. Given analytical method $m$ and its network construction parameters $\boldsymbol{\theta}_{m}$, the corresponding brain network representation can be expressed as

\[
\mathcal{R}_{s,m}=\mathcal{C}_{m}\left(\mathbf{X}_{s},\mathbf{S}_{s};\boldsymbol{\theta}_{m}\right),
\]

where $\mathcal{C}_{m}$ denotes the temporal partitioning, connectivity estimation, and network organization procedures used by method $m$, and $\mathcal{R}_{s,m}$ is the resulting brain network representation.

Most methods that use dynamic functional connectivity as input adopt a sliding-window framework (Lurie et al., 2020). The module first applies signal processing to the ROI time series according to the method settings. It then segments the time series using window length $L$ and actual step size $\Delta$. When the step size is determined by overlap ratio $\eta$, the number of windows is

\[
K=\left\lfloor\frac{M-L}{\Delta}\right\rfloor+1,\qquad \Delta=\max\left\{1,\left\lfloor L(1-\eta)\right\rfloor\right\}.
\]

Assuming that the signals are approximately stationary within each local time interval, let $\mathbf{X}_{s}^{(k)}$ denote the ROI signals in window $k$. The corresponding connectivity matrix is

\[
\mathbf{A}_{s,m}^{(k)}=\mathcal{F}_{m}\left(\mathbf{X}_{s}^{(k)};\boldsymbol{\phi}_{m}\right),\qquad k=1,2,\ldots,K,
\]

where $\mathcal{F}_{m}$ denotes the connectivity estimation procedure used by the method. Each time window therefore yields an $N\times N$ brain network. Arranging all windows chronologically produces a dynamic brain network sequence $\mathcal{A}_{s,m}\in\mathbb{R}^{K\times N\times N}$. After estimating the connectivity matrices, the module applies method-specific operations to obtain the network representations used in subsequent analyses. These operations include processing connection signs, converting weights to absolute values, symmetrization, self-connection removal, and thresholding during network construction. Methods using a fixed number of windows or other specialized temporal partitioning schemes determine window lengths, step sizes, and actual temporal ranges according to their own rules.

After network construction, the module displays the resulting networks. For dynamic networks, users can select a participant and an actual time window. Connectivity matrices can be displayed sequentially to examine changes in interregional connectivity patterns over time. For static networks, when K is set to 1, the module directly displays the corresponding individual or group connectivity matrix. For multimodal methods, it also visualizes structural connectivity.

\subsection{Brain Network Feature Extraction}

Existing tools provide limited integrated support for characterizing dynamic connectivity evolution, complex spatiotemporal dependencies, and higher-order topological relationships. This module therefore emphasizes feature extraction methods that capture connectivity changes across windows and multilevel network organization. It also covers joint structure--function representations and static connectivity analysis. For participant $s$, let the brain network representation under method $m$ be $\mathcal{R}_{s,m}$. Let $\mathcal{U}_{s,m}$ denote synchronized node signals or auxiliary modality information. The corresponding feature extraction process can be expressed as

\[
\mathbf{h}_{s,m}=\mathcal{E}_{m}\left(\mathcal{R}_{s,m},\mathcal{U}_{s,m};\boldsymbol{\theta}_{m}\right),
\]

where $\mathcal{E}_{m}$ denotes the method-specific feature encoding process. The resulting $\mathbf{h}_{s,m}$ may comprise connectivity edge vectors, node embeddings, graph-level vectors, spatiotemporal representations of dynamic graphs, or joint cross-modal representations. Based on the information used during the forward pass, the 27 integrated methods are grouped into four categories. These comprise dynamic temporal representations, complex spatiotemporal and higher-order topological representations, joint structure--function representations, and static association and temporally aggregated topological representations.

Dynamic temporal methods primarily preserve the window order of node signals or connectivity patterns, together with dependencies across windows. GAT (Yang et al., 2019) encodes dynamic functional graphs within each window. It then produces participant-level dynamic graph representations through temporal attention, BiGRU, or mean aggregation. STGCN (Gadgil et al., 2020) jointly extracts interregional propagation and local temporal patterns from blood oxygen level-dependent (BOLD) subsequences, constrained by a static functional topology shared across participants. STAGIN (Kim et al., 2021) uses GIN to extract topological representations for individual windows and applies a node-attention readout to obtain window-level graph vectors. A Transformer then models dependencies among dynamic functional graphs. CNN, BiLSTM, and Transformer (Meszl{\'{e}}nyi et al., 2017; Yan et al., 2018; Kan et al., 2022) extract local temporal patterns, bidirectional window context for each brain region, and global dependencies across windows, respectively, from windowed ROI signals. For these three baselines, the citations indicate related applications of the model families in brain network analysis, while the toolkit uses generic implementations for windowed ROI signals; its BiLSTM baseline is distinct from the Full-BiLSTM architecture of Yan et al. (2018). Together, these methods capture local temporal changes, dependencies across windows, and evolving functional topology at the levels of both node activity and connectivity patterns.

Building on dynamic temporal representations, complex spatiotemporal and higher-order topological methods further characterize functional subnetworks, window--region interactions, and hypergraph relationships. ST2GCN (Li et al., 2025c) partitions dynamic functional graphs into potentially overlapping functional subnetworks. It jointly extracts within-window topological features within subnetworks and propagation features across windows for the same brain region. CD-DSTCN (Yuan et al., 2025) constructs Pearson connectivity graphs internally from segmented BOLD signals. It integrates node representations from different temporal segments through temporal attention, spatiotemporal convolution, and confidence-weighted fusion. LR-STIGCN (Li et al., 2025b) jointly encodes within-window functional connectivity, window--region interaction graphs, and learnable hypergraph relationships to generate higher-order node representations. Constrained by dynamic functional connectivity, NeuroH-TGL (Li et al., 2025d) separates spatial components that remain consistent across windows from time-varying components. It then derives heterogeneous propagation weights from relationships across windows. STHAN (Li et al., 2026) combines DTW temporal priors with an internally constructed KNN--Louvain hypergraph to jointly extract temporal and spatial attention features. These methods extend feature representations from individual window graphs to functional subnetworks, joint graphs spanning windows, heterogeneous propagation relationships, and hypergraph structures. They capture complex spatiotemporal topological features, including multiscale functional modules, higher-order window--region interactions, heterogeneous propagation across windows, and coordination among multiple brain regions.

Joint structure--function methods further use diffusion tensor imaging (DTI) connectivity to constrain or complement dynamic functional features. OT-MCSTGCN (Zhu et al., 2024) constructs higher-order representations of topological evolution using PageRank hub distributions from dynamic functional graphs and DTI-derived transport costs. It also constrains graph convolution using DTI-derived Chebyshev bases, thereby extracting dynamic functional features under structural constraints. MSTGAC (Li et al., 2025a) uses rows of dynamic functional connectivity matrices as node features and nonzero DTI connections as graph attention masks. It combines multiview consistency constraints with LSTM to form structurally constrained representations of functional temporal dynamics. ADRNet (Cui et al., 2026) encodes fMRI and DTI data into modality vectors, forming joint representations through cross-modal generation and missing-modality reconstruction. These multimodal methods capture structural constraints on functional propagation and temporal evolution, structure--function coupling, and complementary cross-modal features.

The module also integrates static association and temporally aggregated topological methods to support conventional static analyses and comparisons with dynamic spatiotemporal representations. Logistic Regression, SVM, Random Forest (Shevchenko et al., 2025), XGBoost (Wu et al., 2023), LDA, CCA--LDA (Lin et al., 2020), and MLP (Kim et al., 2016) use connectivity vectors constructed in one of two ways. The first extracts the strictly upper-triangular entries after averaging functional connectivity across windows. The second concatenates upper-triangular edge weights from individual windows in temporal order. These methods characterize linear connectivity contributions, maximum-margin information, nonlinear threshold interactions, supervised discriminant directions, canonical components associated with class encodings, and global nonlinear combinations. Specifically, CCA--LDA computes canonical components from connectivity edge vectors and class encodings; it does not model associations across imaging modalities. For temporally aggregated graph topology, BrainNetCNN (Kawahara et al., 2017) extracts connectome features hierarchically across edges, nodes, and the whole graph. GCN (Ktena et al., 2017), GraphSAGE (Wang et al., 2022), and GIN (Kim and Ye, 2020) derive node representations from multihop neighborhoods, weighted full neighborhoods, and varying propagation depths, respectively. BrainGNN (Li et al., 2021) uses ROI connectivity profiles as node features and applies ROI-aware convolution and hierarchical pooling to extract discriminative brain regions and subgraphs. SimSiam (Wang et al., 2025) learns graph-level representations from aggregated brain graphs that remain stable under node and edge perturbations. Within the module, these methods extract static connectivity, discriminative edge patterns, and aggregated topological features, supporting conventional brain network analysis and classification.

Overall, the module provides a multilevel feature extraction framework spanning dynamic temporal, complex spatiotemporal, higher-order topological, and joint structure--function representations. It also accommodates static connectivity and temporally aggregated topology analyses. This framework offers comprehensive methodological support for analytical tasks across different brain network representations.

\subsection{Brain Biomarker Discovery}

The high-dimensional representations produced by brain network models are difficult to relate directly to specific brain regions and connections. This module therefore extracts internal model signals after training and maps them into a common representation of group-level connection and regional importance. Assuming that the current analysis sample includes $S$ participants, the interpretation procedure for method $m$ can be expressed as

\[
\left(\mathbf{W}_{m},\mathbf{r}_{m}\right)=\mathcal{I}_{m}\left(\left\{\mathcal{R}_{s,m},\mathcal{U}_{s,m},\mathbf{h}_{s,m},y_s\right\}_{s=1}^{S};\widehat{\boldsymbol{\theta}}_{m}\right),
\]

where $\mathcal{I}_{m}$ denotes the method-specific interpretation procedure and $\widehat{\boldsymbol{\theta}}_{m}$ denotes the available trained model parameters. The quantities $\mathbf{W}_{m}\in\mathbb{R}_{\geq0}^{N\times N}$ and $\mathbf{r}_{m}\in\mathbb{R}_{\geq0}^{N}$ represent connection importance and regional importance, respectively. The biomarker discovery procedures for each feature extraction category are described below.

For dynamic temporal methods, GAT combines edge attention with connectivity proxy coefficients and can provide time-window weights under specific aggregation modes. STGCN combines individual functional connectivity proxies with a learnable adjacency modulation matrix. STAGIN maps temporally aggregated node attention to brain regions and constructs derived connections through the outer product of node scores. CNN, BiLSTM, and Transformer primarily use intermediate temporal representations to support functional connectivity proxy analysis.

For complex spatiotemporal and higher-order topological methods, ST2GCN projects subnetwork attention back to the constituent brain regions and region pairs. CD-DSTCN uses its internal dynamic Pearson adjacency matrices to represent connection importance and computes regional importance from the norms of the final node features. LR-STIGCN combines connectivity proxies with node representation scores. NeuroH-TGL uses STPD representations to support connectivity proxy analysis. STHAN combines spatial attention signals extracted by the model with functional connectivity proxies. These approaches generate rankings of candidate connections and brain regions at different organizational scales.

For joint structure--function methods, OT-MCSTGCN constructs a discriminative proxy from higher-order topological representations. MSTGAC currently derives importance rankings from window-specific functional connectivity. ADRNet evaluates sensitivity to masking ROIs and DTI connections and incorporates functional and structural connectivity proxy signals. This characterizes the influence of multimodal information on model outputs.

For static association and temporal aggregation methods, Logistic Regression and SVM derive connection importance from linear coefficients and kernel model approximations, respectively. Random Forest and XGBoost use tree impurity reduction and split gain, respectively. LDA and CCA--LDA use discriminant and canonical loadings, respectively. XGBoost additionally incorporates SHAP values when applicable (Lundberg and Lee, 2017). MLP uses connectivity proxies supported by hidden representations. BrainNetCNN uses intermediate E2E and E2N activations to characterize edge and node signals. GCN, GraphSAGE, GIN, and SimSiam use connectivity proxies supported by model representations. BrainGNN combines connectivity proxies with TopK pooling gate scores.

The final outputs are organized into important brain connections and regions, presented as connectivity matrices, Top-K connections, and regional rankings. The module supports matrix displays and two- and three-dimensional brain maps, with filtering by threshold, brain lobe, and functional network. Results can also be exported in CSV format. These importance scores describe the relative discriminative strength of connections and regions within the current sample. They represent method-specific, group-level candidate biomarker screening results and key brain network patterns, providing a basis for further investigations of neural mechanisms.

\subsection{LLM Reports Based on Structured Evidence}

To translate brain network analysis results into verifiable statistical evidence, this module takes ROI-level fMRI time series and preconstructed DTI structural connectivity matrices as input. It constructs functional connectivity (FC) and structural connectivity (SC) networks and extracts features at the edge, node, global, and functional system levels. The module then performs individual- or group-level statistical analyses and encodes the results as structured evidence for generating supporting analytical reports.

For fMRI data, let the ROI-level BOLD time series of participant $s$ be denoted by $\mathbf{X}_{s}\in\mathbb{R}^{N\times T_s}$. Interregional Pearson correlations are calculated using sliding windows of length $L$ and step size $\Delta$. The correlations are Fisher-transformed after replacing undefined values with zero and averaged across windows. A weighted, undirected functional network is then constructed using the threshold $\tau_{\mathrm{FC}}=0.1$:

\[
\begin{aligned}
K_s&=\max\left\{1,\left\lfloor\frac{T_s-L}{\Delta}\right\rfloor+1\right\},\\[4pt]
\bar{z}_{s,ij}&=\frac{1}{K_s}\sum_{k=1}^{K_s}\operatorname{arctanh}\!\Bigl[\operatorname{clip}\!\Bigl(\\[-2pt]
&\qquad\operatorname{corr}\!\left(X_{s,i,\mathcal{W}_k},X_{s,j,\mathcal{W}_k}\right),-1+\varepsilon,1-\varepsilon\Bigr)\Bigr],\quad i\ne j,\\[4pt]
A_{s,ij}^{\mathrm{FC}}&=\begin{cases}
|\bar{z}_{s,ij}|,& |\bar{z}_{s,ij}|\geq\tau_{\mathrm{FC}},\ i\ne j,\\
0,&\text{otherwise}.
\end{cases}
\end{aligned}
\]

Here, $\varepsilon=10^{-4}$ and $\overline{\mathbf{Z}}_s=(\bar{z}_{s,ij})$, with $\bar{z}_{s,ii}=0$. For $T_s<L$, all time points form a single window. Mean FC, mean absolute FC, the proportion of positive connections, and connection density are calculated from the temporally averaged functional connectivity. Dynamic functional connectivity measures at whole-brain and subnetwork levels are also derived from the standard deviation and coefficient of variation of connectivity across windows.

For DTI data, the module takes the structural connectivity matrix $\mathbf{S}_s^{\mathrm{raw}}$, obtained during diffusion data preprocessing and tractography, as input. The matrix is symmetrized, its diagonal is set to zero, and its values are normalized by the maximum absolute value within each participant. Positive connections meeting the default threshold $\tau_{\mathrm{SC}}=0.05$ are retained:

\[
\begin{aligned}
\mathbf{S}_{s}^{\mathrm{sym}}&=\frac{\mathbf{S}_{s}^{\mathrm{raw}}+\left(\mathbf{S}_{s}^{\mathrm{raw}}\right)^{\mathsf{T}}}{2},\qquad S_{s,ii}^{\mathrm{sym}}=0,\\[4pt]
m_s&=\max_{i\ne j}\left|S_{s,ij}^{\mathrm{sym}}\right|,\\[4pt]
A_{s,ij}^{\mathrm{SC}}&=\begin{cases}
S_{s,ij}^{\mathrm{sym}}/m_s,&m_s>0,\ S_{s,ij}^{\mathrm{sym}}/m_s\geq\tau_{\mathrm{SC}},\ i\ne j,\\
0,&\text{otherwise}.
\end{cases}
\end{aligned}
\]

The resulting structural network is used to calculate mean relative weight across all possible edges, connection density, the number of retained connections, and maximum connection weight. The mean and standard deviation of nonzero edge weights are also calculated to characterize the extent of structural connectivity and its weight distribution.

After both network types are obtained, a unified graph-theoretical framework extracts topological features at the edge, node, and global levels (Rubinov and Sporns, 2010). Node-level measures include node strength, weighted degree, clustering coefficient, betweenness centrality, and nodal local efficiency. Hub nodes are identified based on weighted degree. FC node strength is calculated by aggregating unthresholded mean absolute functional connectivity, whereas SC node strength aggregates thresholded relative structural connectivity. Global measures include global efficiency, local efficiency, characteristic path length, mean clustering coefficient, modularity, small-world coefficient, rich-club coefficient, and assortativity coefficient. Efficiency measures and characteristic path length are calculated using topological distances in the thresholded networks. Connection weights are retained when calculating clustering coefficients and modularity.

In addition to the unified graph-theoretical measures, analyses are performed at the functional system level. The brain parcellation is mapped to eight potentially overlapping subnetworks. These comprise the default mode network (DMN), frontoparietal network (FPN), salience network (SAN), attention network (ATN), sensorimotor network (SMN), visual network (VN), auditory network (AN), and olfactory/orbitofrontal network (ON). Let $\mathcal{R}_a$ denote the set of brain regions in subnetwork $a$, and define $B_s^{\mathrm{FC}}=|\overline{\mathbf{Z}}_s|$ and $B_s^{\mathrm{SC}}=\mathbf{A}_s^{\mathrm{SC}}$. System-level connectivity for modality $m$ is then expressed as

\[
C_{s,ab}^{(m)}=\begin{cases}
\displaystyle\frac{2}{n_a(n_a-1)}\sum_{\substack{i<j\\i,j\in\mathcal{R}_a}}B_{s,ij}^{(m)},&a=b,\\[8pt]
\displaystyle\frac{1}{n_a n_b}\sum_{i\in\mathcal{R}_a}\sum_{j\in\mathcal{R}_b}B_{s,ij}^{(m)},&a<b,
\end{cases}\qquad m\in\{\mathrm{FC},\mathrm{SC}\},
\]

where $n_a=|\mathcal{R}_a|$. Between-network means include all entries of the cross-network submatrix, including zero-valued self-connection entries ($i=j$) for shared ROIs. This yields eight within-network and 28 between-network mean connectivity measures for each modality. When both fMRI and DTI data are available, structure--function coupling is additionally calculated at the global, regional, and subnetwork levels (Fotiadis et al., 2024). At the individual level, correlations between functional connectivity in each time window and structural connectivity describe changes in coupling. Connectivity strength quadrants and dissociated connections are also combined to form joint cross-modal representations.

Statistical analyses are conducted at both individual and group levels. At the individual level, a designated reference group is used to estimate distributions of global and regional measures and calculate standardized deviations. Measures meeting predefined non-normality criteria undergo a Blom rank-based inverse normal transformation (McCaw et al., 2020). When no reference group is available, global measures are evaluated against built-in reference distributions from the literature. Regional measures are described using within-participant rank percentiles, without inferring abnormalities across participants. For regional analyses with a reference group, all valid region-by-measure combinations are included in a single family of tests. Benjamini--Hochberg false discovery rate correction is applied before Top-K selection (Benjamini and Hochberg, 1995). When the individual belongs to the reference group, the squared Mahalanobis distance of the global measure vector further characterizes their overall deviation. At the group level, Welch's two-sample $t$ tests are applied to both global and regional measures. Cohen's $d$ quantifies standardized between-group differences. Significance inference for global measures uses $p$ values without multiple-comparison correction. Regional measures undergo Benjamini--Hochberg correction across the complete family of tests. All group comparisons are performed without covariate adjustment. Once the sample size reaches a predefined minimum, network-based statistics (NBS) are applied separately to structural and functional connectivity (Zalesky et al., 2010). Connections are selected using an edge-level threshold $p_0$, and connected components are constructed within a two-sided testing framework. Direct permutation of group labels is performed $B$ times to obtain the null distribution of the maximum component edge count. This controls the family-wise error rate at the component level.

Statistical results are subsequently organized into structured summaries containing data modality, analysis level, measure values, statistical tests, effect direction, effect size, multiple-comparison correction status, and evidence sources (Gao et al., 2023). The system prompt specifies the levels of statistical evidence interpretation and the report fields. Depending on the analysis mode, the user prompt conveys global measures, participant-level assessments, Top-$K$ regional results, NBS components, representative connections, structure--function coupling, and group comparisons. Complete connectivity matrices are not transmitted. The LLM uses this information to generate a candidate report in a predefined JSON structure. The backend reconstructs the statistics and narrative for formal group differences from the source data. It also applies rule-based validation and tiered presentation to other verifiable fields. The final structured supporting report integrates global features, regional results, significant connectivity components, cross-modal features, and group differences.

\section{Toolkit Analysis Workflows}

In practical applications, the four functional modules described in Section 2 form two relatively independent brain network analysis workflows. The first sequentially connects brain network construction and visualization, brain network feature extraction, and brain biomarker discovery. It begins with method selection and data import and proceeds through brain network modeling, feature learning, and the identification and visualization of brain biomarkers. The second centers on the LLM reporting module based on structured evidence. It accepts processed unimodal or multimodal data, constructs brain networks, and calculates metrics at the individual or group level. The results are organized into structured evidence, which the LLM synthesizes and interprets to generate a brain network analysis report. From a practical application perspective, this section describes the input data, key processing steps, and outputs of both workflows.

\subsection{Brain Network Construction and Visualization, Feature Extraction, and Candidate Brain Biomarker Discovery}

Studies of existing brain network and graph learning toolkits indicate that, in addition to integrating algorithms, platforms for practical research should organize network construction, task definition, model training, and result interpretation into a continuous, traceable workflow (Zhou et al., 2020; Waller et al., 2018). The toolkit uses analytical method selection as the entry point to the workflow. The method selection page summarizes the data requirements, network representations, supported tasks, and output types for each method. A method adaptation mechanism automatically links the corresponding data validation, brain network construction, feature extraction, training, and interpretation functions. Using ST2GCN (Li et al., 2025c) as an example, this section presents the complete workflow, from method selection, brain network construction, and parameter configuration to feature extraction, classification, and candidate brain biomarker discovery.

During brain network construction, users first upload neuroimaging data that meet the requirements of the selected method. After parsing the files, the toolkit displays the data structure, sample count, and label information for verification. Network construction parameters depend on the selected feature extraction method and can be configured through a graphical form. Based on these settings, the toolkit preprocesses the ROI-level time series, partitions them into time intervals, and estimates connectivity to generate the corresponding brain network representations. The resulting networks are visualized as connectivity matrices and circular network plots. Connectivity matrices allow users to inspect overall connection strengths and the distribution of network modules, while circular plots highlight connections between specific ROIs and their changes across windows. Users can select a participant, input layer, and time window, then examine network changes over time using a slider, forward and backward navigation, or automatic playback. The interface supports ROI filtering by brain lobe or functional network. Selections within the same filtering dimension are combined by union, whereas selections across dimensions are combined by intersection. Filters remain consistent when switching between windows. Visualization settings affect only the display and do not alter brain network construction. The page also displays the brain atlas, preprocessing settings, window ranges, connectivity estimation method, and construction threshold. This information allows users to verify the dynamic networks and their construction parameters while exploring an intuitive, interactive visualization. Figure 3-1 shows the networks covering all ROIs that serve as ST2GCN model inputs for the same participant across four consecutive time windows.

\begin{figure}[!htbp]
\centering
\includegraphics[width=\linewidth,height=0.62\textheight,keepaspectratio]{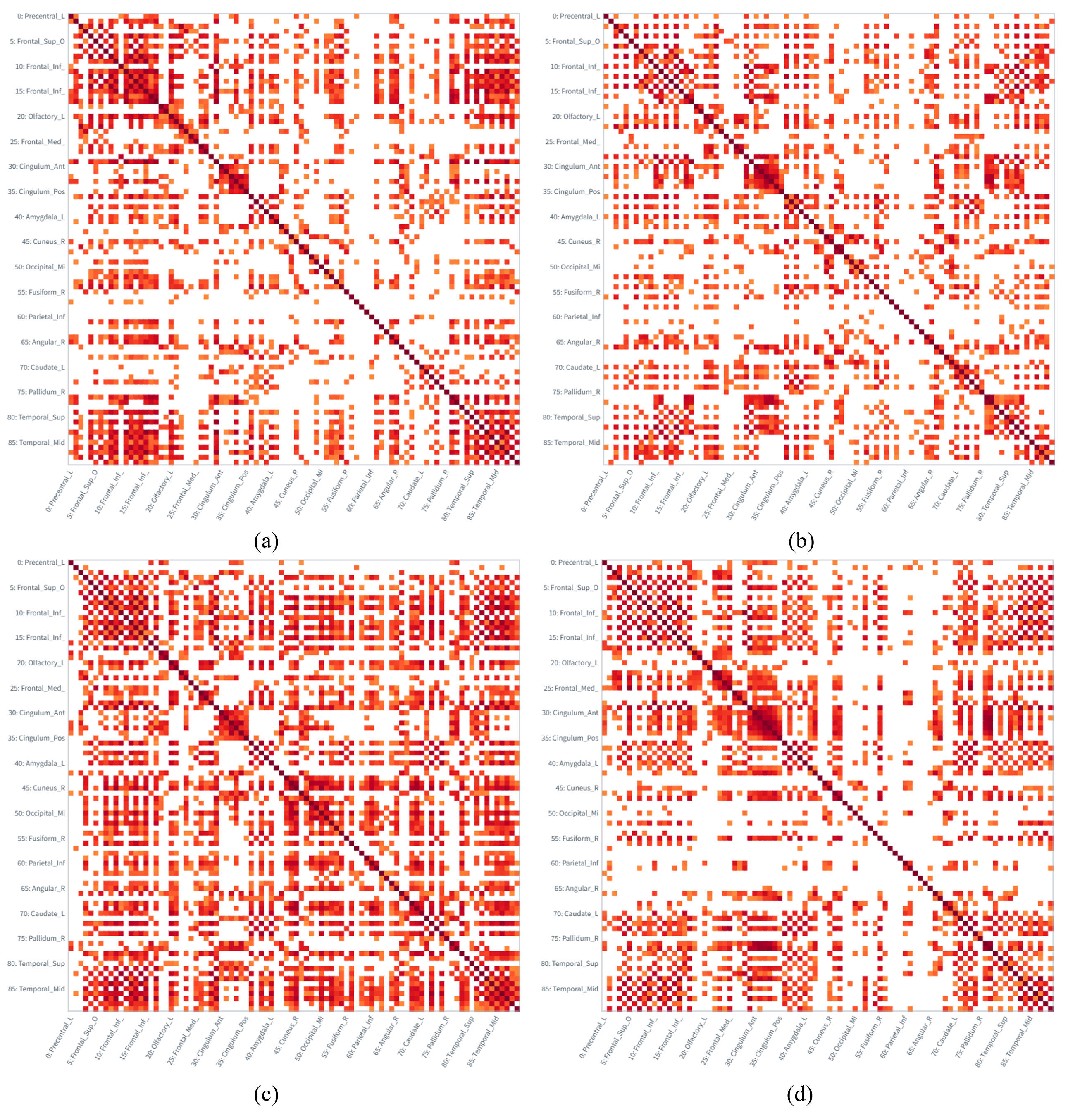}
\caption{Dynamic brain network input matrices for ST2GCN. Panels (a), (b), (c), and (d) show connectivity matrices covering all ROIs in four consecutive time windows from the same participant. Identical ROI ordering and display settings are used to illustrate changes in brain connectivity patterns over time.}
\label{fig:3-1}
\end{figure}

After dynamic brain network construction, the toolkit proceeds to brain network feature extraction. The workflow comprises data parameter configuration, model parameter configuration, training parameter configuration, feature extraction, and downstream tasks. During data parameter configuration, the toolkit automatically reads the original labels, counts the samples associated with each label, and displays the class distribution. A graphical task builder supports mapping one or more original labels to a target group. Samples not assigned to a target group are excluded from the current task. As the configuration changes, the task preview updates the target classes, included and excluded samples, and class distribution. Users can therefore identify label mapping errors, missing samples, or class imbalance before training. Data organization settings, including the training data proportion and the number of cross-validation folds, are saved with the task definition. This step establishes a consistent correspondence among network samples, original labels, and target classes and organizes the networks prepared for the selected method into inputs for a specific downstream task.

Once the task is defined, the toolkit provides graphical configuration of model and training parameters. The model parameter page dynamically generates a configuration form for the selected method. For ST2GCN, the form includes architectural parameters such as window settings, hidden feature dimensions, and dropout. The training parameter page brings together optimization settings, including batch size, training epochs, learning rate, weight decay, and early stopping criteria. The toolkit also supports grid search, allowing users to specify candidate values for multiple hyperparameters, together with the number of cross-validation folds and the maximum number of epochs. The resulting search space can be saved with the task and reviewed later. These configurations organize the architectural constraints and optimization strategies of different methods into structured experimental settings, providing a common input for subsequent feature extraction and training coordination.

After configuration, the toolkit coordinates model training, cross-validation, run monitoring, and result archiving to perform brain network feature extraction and downstream tasks. Dynamic brain networks are passed to the model through method-specific interfaces. The model learns brain network features and uses them for classification. The training procedure follows the specified cross-validation scheme for parameter updates, early stopping, and model checkpoint saving. After training, the toolkit summarizes classification performance across folds and saves the task configuration, metrics for each fold, and model files. It then automatically invokes the result precomputation routine associated with the selected method, organizing model-derived interpretation signals into important brain connections and regions. This completes the continuous workflow from brain network input, feature learning, and downstream tasks to brain biomarker discovery.

\begingroup
\setlength{\intextsep}{6pt}
\begin{figure}[!htbp]
\centering
\captionsetup{hypcap=false,skip=4pt}
\includegraphics[width=0.90\linewidth,height=0.72\textheight,keepaspectratio]{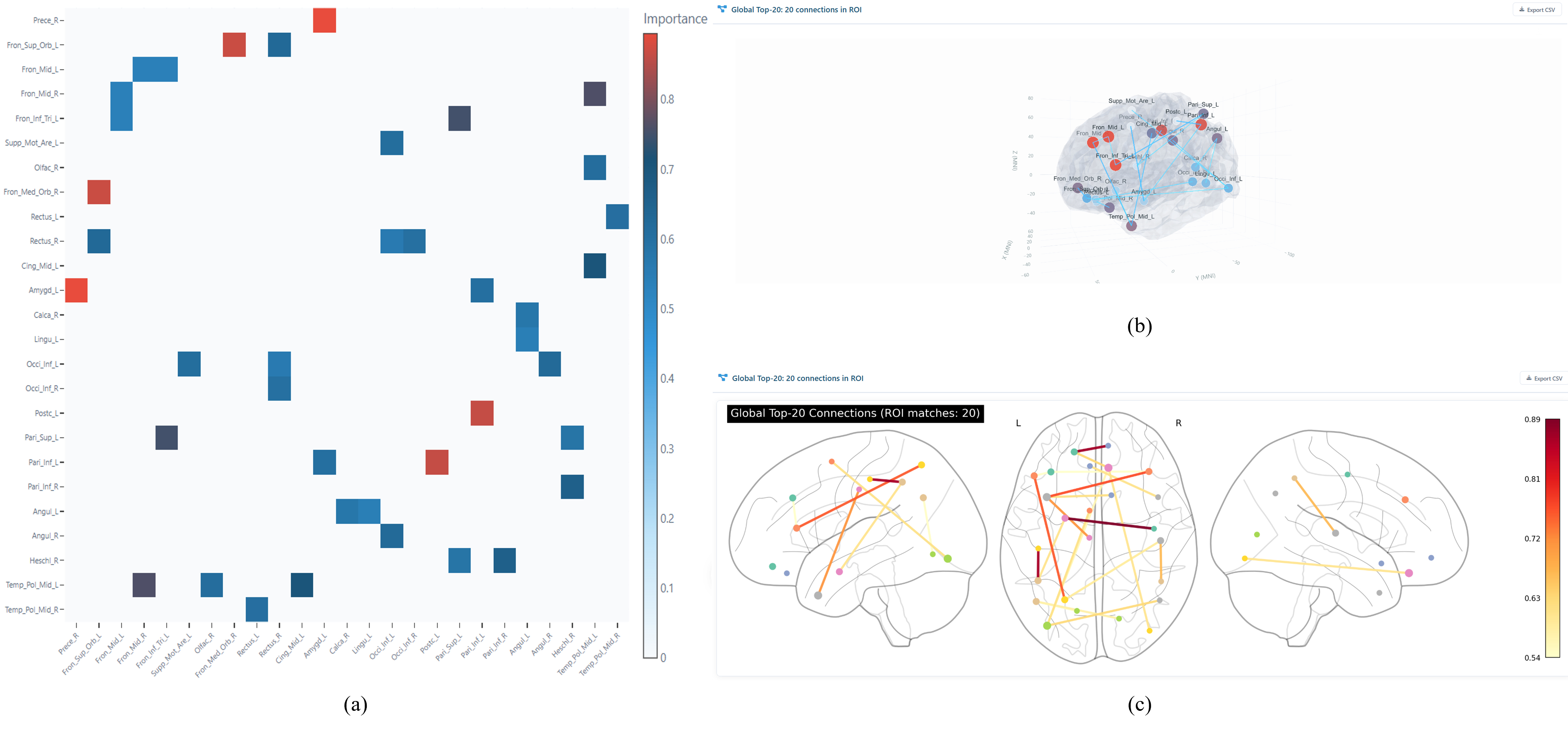}
\caption{ST2GCN-derived important brain connections. (a) Connectivity matrix of the global Top-20 connections. (b) Three-dimensional view with nodes at MNI coordinates; line width and opacity indicate relative connection importance. (c) Left lateral, axial, and right lateral projections of the same connections.}
\label{fig:3-2}
\par\vspace{4pt}
\includegraphics[width=0.90\linewidth,height=0.72\textheight,keepaspectratio]{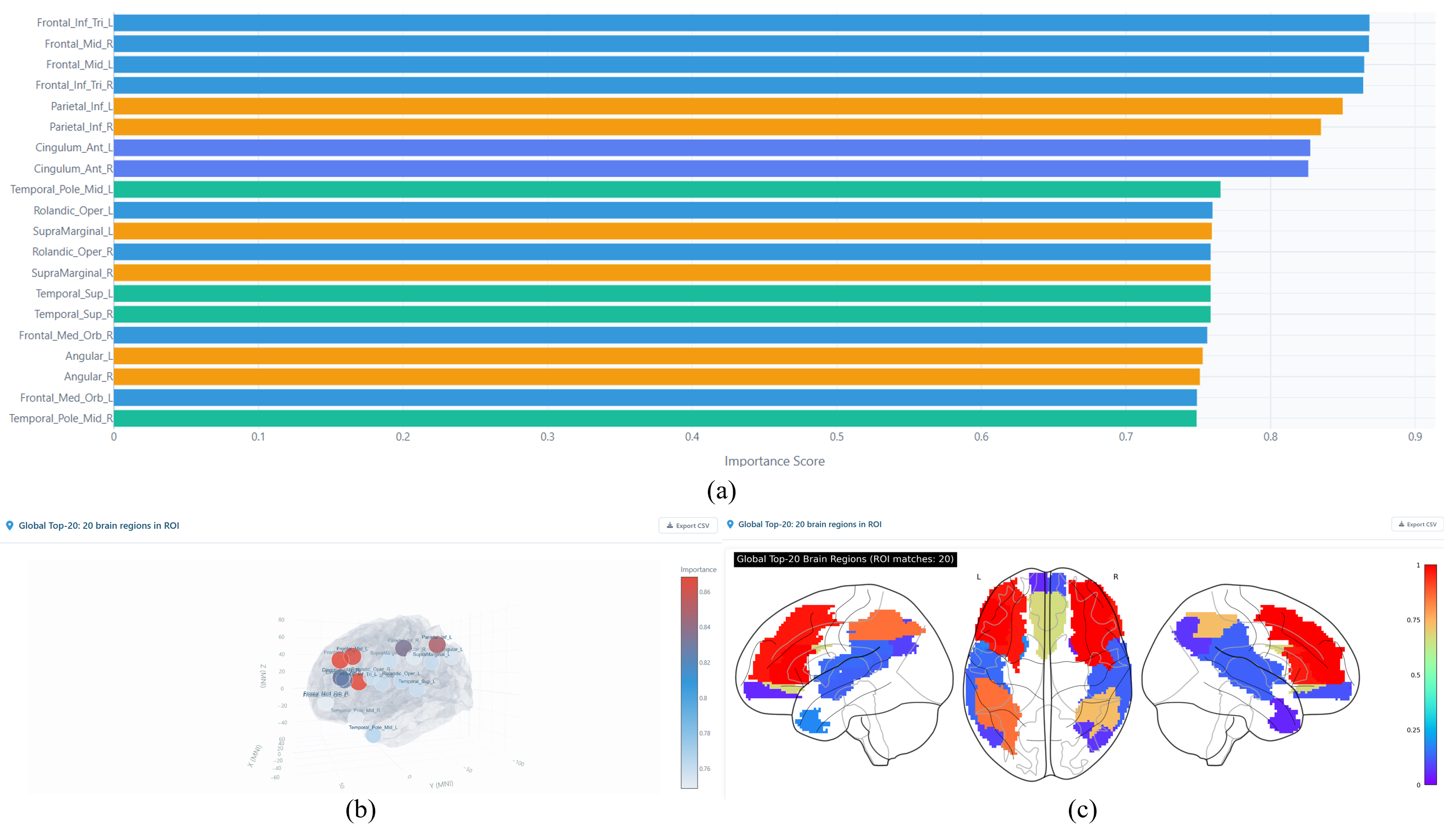}
\caption{ST2GCN-derived important brain regions. (a) Global Top-20 regions, with bar length showing normalized importance. Colors denote AAL90 anatomical groups (Tzourio-Mazoyer et al., 2002): frontal (blue), parietal (orange), cingulate (blue-violet), and temporal (turquoise). (b) Three-dimensional view; node size and color indicate relative regional importance. (c) Left lateral, axial, and right lateral projections of the same regions.}
\label{fig:3-3}
\end{figure}
\endgroup

After brain biomarker discovery, the toolkit presents results at both the connection and regional levels. Important connections are displayed as connectivity matrices, three-dimensional brain maps, and two-dimensional views from three perspectives. Important regions are displayed as bar charts, three-dimensional brain maps, and corresponding two-dimensional views. In the three-dimensional maps, nodes are positioned using MNI coordinates. Node size and color indicate regional importance, while line width and opacity indicate connection importance. Users can rotate and zoom the maps and inspect nodes by hovering over them. The two-dimensional connection views show the same connections from left lateral, axial, and right lateral perspectives. The regional views render filled atlas voxels to show the spatial distribution of important regions from the same perspectives. The toolkit also supports ROI filtering by brain lobe or functional network to restrict the analysis scope. For both connections and regions, the global Top-K results are selected before the current ROI filter is applied. Connections are retained only when both endpoints belong to the selected ROI set, and regions are retained only if they belong to that set. Figures 3-2 and 3-3 show the important connections and regions identified by the example method and their spatial distributions.

\FloatBarrier

\subsection{LLM-Assisted Brain Network Analysis Reports Based on Structured Evidence}

The LLM reporting module based on structured evidence supports individual and group analyses. Both modes follow the same overall workflow: LLM configuration, data import, selection of the participant or groups to be analyzed, result verification, and report generation. They differ primarily in the analysis targets and levels of evidence. Individual analysis focuses on the brain network characteristics of a selected participant and their deviations from reference values, whereas group analysis examines global, regional, and connectivity differences between a reference group and a target group. To illustrate how these stages are connected, this section presents a group analysis using multimodal data, covering the complete workflow from task setup and result inspection to LLM-assisted report generation.

On the analysis page, users first configure the LLM service and model, then import processed fMRI, DTI, or multimodal data. After loading the data, the toolkit provides an overview of the input modalities, numbers of participants and brain regions, and label distribution, allowing users to verify the data included in the analysis. For individual analysis, users select a target participant from the valid data. For group analysis, they specify the reference and target groups to define the subsequent comparison. Brain network analysis can begin after the analysis mode, selected participant or groups, and related settings have been verified. Individual and group analyses thus share a common configuration interface before proceeding to result inspection and report generation.

After group analysis, the toolkit first provides a group overview summarizing the reference and target groups actually included, the available data modalities, and the results generated. Users can confirm the scope of the comparison and identify the types of evidence available for inspection. Following this initial verification, the group comparison page displays brain network metrics for the two groups alongside effect sizes, allowing users to examine differences in global network properties in terms of both direction and magnitude. Because global metrics alone cannot localize these differences, regional results are examined next. Between-group tests and multiple-comparison correction identify differences in specific ROIs, with changes in regional metrics and the associated functional networks displayed together. This helps users identify brain regions and network systems in which differences are concentrated. When the relevant analysis requirements are met, users can also inspect network-level statistical results and group-averaged structural connectivity (SC) and functional connectivity (FC) matrices to examine pairwise interregional connections and overall connectivity changes. Global brain network metrics and regional results therefore provide the core evidence, progressing from overall characteristics to local spatial patterns. Figure 3-4 presents group analysis results at these two levels.

\begin{figure}[!htbp]
\centering
\includegraphics[width=\linewidth,height=0.72\textheight,keepaspectratio]{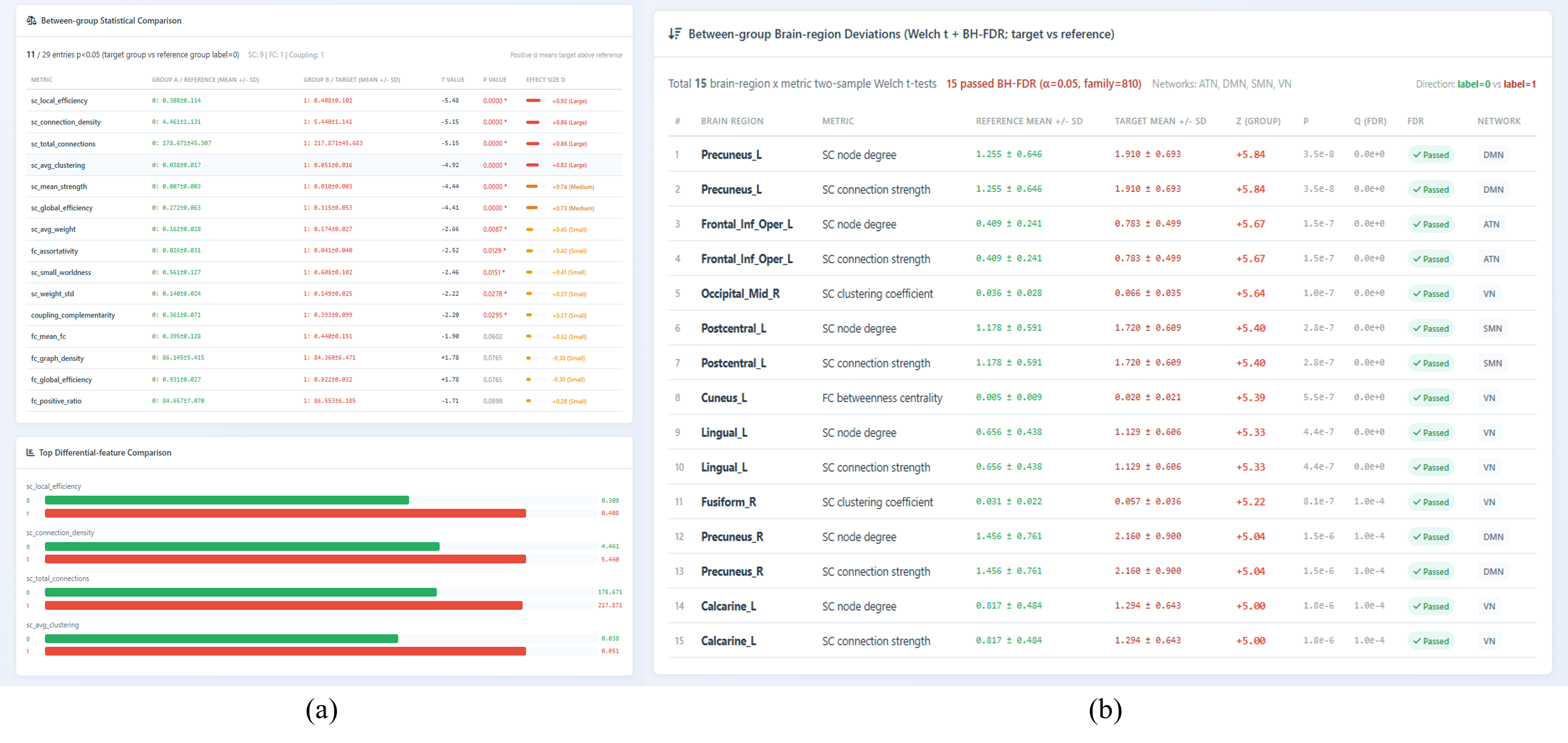}
\caption{Global and regional evidence of between-group differences in brain networks. (a) Comparisons of global brain network metrics and corresponding effect sizes between the reference group (n = 73) and the target group (n = 70), including a measure of structure--function coupling complementarity. (b) ROIs showing significant between-group differences after Welch's tests and Benjamini--Hochberg false discovery rate (BH-FDR) correction (Benjamini and Hochberg, 1995), together with changes in their metrics and associated functional networks.}
\label{fig:3-4}
\end{figure}

After the results at each level have been verified, the toolkit organizes the data modalities, group definitions, between-group metrics, regional results, significant connectivity components, and available cross-modal information into a structured feature summary. Before generating a report, users can inspect the summary to be submitted to the LLM and add background information relevant to the analysis targets according to their research needs. This step allows the scope of the data and the main evidence underlying the report to be confirmed before generation. The LLM then synthesizes and organizes the existing results into a comprehensive narrative interpretation.

Once generated, the report presents the group analysis results as both a structured summary and a complete narrative. The structured section displays an overview, relevant networks, and major between-group differences, together with significant connectivity components and cross-modal results when supported by the available evidence. The complete narrative integrates evidence across levels, progressing from global to local findings. Metrics and differences presented in the report correspond to the preceding analysis results, and users can return to the relevant pages to verify them. Reports and analysis results can also be exported in Markdown, JSON, or CSV format for subsequent organization and review.

\section{Toolkit Application and Evaluation}

Using ADNI data, this section demonstrates the toolkit's capabilities in brain network construction and visualization, brain network feature extraction, and brain biomarker discovery. First, ST2GCN is used as an example to present the constructed brain networks and their visualization, illustrating the organization and connectivity patterns of the network inputs. Next, the classification performance of the 27 integrated brain network feature extraction methods is evaluated. Finally, the important brain regions and connections identified by ST2GCN are examined to demonstrate the spatial distribution and analysis of candidate brain biomarkers. Following the sequence of brain network construction, classification evaluation, and brain biomarker discovery, these experiments assess the toolkit's support for practical neuroimaging analysis and provide a reference for researchers conducting brain network studies.

\subsection{ADNI Data, Evaluation Protocol, and Parameter Sources}

This section uses paired fMRI and DTI data from ADNI (Weiner et al., 2025), comprising 203 participants: 73 cognitively normal controls (NC), 70 participants with Alzheimer's disease (AD), and 60 participants with other cognitive statuses. The primary supervised learning task is binary classification between NC and AD, including 143 participants. The remaining 60 participants are excluded from this task. The two classes are similar in size, providing a basis for examining the discriminative performance of different brain network analysis methods in the current cohort.

The toolkit accepts preprocessed ROI time series and structural connectivity matrices as standardized inputs. In the complete dataset, the fMRI data have dimensions of 203 \(\times\) 90 \(\times\) 197, corresponding to the number of participants, the 90 brain regions defined by the AAL atlas (Tzourio-Mazoyer et al., 2002), and 197 time points, respectively. The DTI data have dimensions of 203 \(\times\) 90 \(\times\) 90 and represent structural connectivity matrices constructed using the same brain parcellation for each participant. Participant correspondence and brain region indices are consistent across the two modalities, providing paired inputs for unimodal analysis and joint structure--function modeling.

The experiment applies the network construction strategy required by each brain network feature extraction method. For fMRI data, the toolkit takes preprocessed ROI time series as input and performs signal processing, temporal partitioning, and interregional connectivity estimation according to the selected method's network construction settings. Connection weights are processed and networks are organized as required to construct static networks, dynamic brain network sequences, or multilayer networks. For methods that use structural information, DTI data are supplied as preconstructed structural connectivity matrices and organized into structural networks or joint structure--function inputs according to the modeling requirements. The different network representations share consistent participant correspondence, brain region indices, and atlas annotations, providing a foundation for subsequent feature extraction, classification evaluation, and spatial mapping of results.

Based on this data organization, classification experiments use 10-fold cross-validation at the participant level. Within each fold, all imaging modalities, time windows, and derived brain network representations from a given participant are assigned to the same subset, preventing data from that participant from appearing in both the training and test sets (Scheinost et al., 2019). Model training, validation, inference, and performance aggregation are managed within the toolkit's task management framework. Data splits, label mappings, random seeds, and model parameters are saved with the task configuration, allowing the experimental conditions and evaluation results for each method to be traced.

Classification performance is evaluated using accuracy (ACC), the area under the receiver operating characteristic curve (AUC), sensitivity, specificity, and the F1 score. ACC measures overall classification accuracy, whereas AUC characterizes the classifier's discrimination across decision thresholds. Sensitivity and specificity reflect the identification of participants with AD and NC participants, respectively, while the F1 score combines precision and recall. Each metric is calculated on the test set of each fold and summarized as the mean \(\pm\) standard deviation across the 10 folds, describing both average classification performance and variation across folds.

\subsection{Brain Network Construction and Visualization Results}

This section uses ST2GCN (Li et al., 2025c) to illustrate the network representations constructed from ADNI data and their visualization. For each participant, the toolkit takes fMRI time series comprising 90 ROIs and 197 time points as input and applies a single Min--Max normalization jointly across all ROIs and time points within that participant. The time series are divided into a fixed number of four non-overlapping windows, each containing 49 time points. These windows cover time points 1--49, 50--98, 99--147, and 148--196, respectively. Because the window length is determined by integer division, the final time point, 197, is excluded. Each participant therefore contributes four temporally ordered sets of ROI signals for connectivity estimation within individual windows.

Within each window, the toolkit calculates Pearson correlation coefficients between brain regions, takes their absolute values, and applies a threshold of 0.6. Connections below the threshold are set to zero, while the remaining connection weights and diagonal self-connections are retained. The resulting four 90 \(\times\) 90 nonnegative weighted connectivity matrices form a dynamic functional network sequence. Figure 4-1 shows the networks constructed for the same participant across four consecutive, non-overlapping windows. The rows and columns of each matrix correspond to the 90 AAL brain regions, and all matrices use the same ROI ordering and color scale to show the connectivity distribution after thresholding. Some areas with relatively concentrated connections persist across multiple windows, while the locations and weights of retained connections vary locally between windows. This allows changes in individual brain region pairs to be examined across windows. Because the network construction procedure uses absolute correlation coefficients, the displayed weights represent correlation strength, and both positive and negative correlations may contribute to the retained connections.

\begin{figure}[!htbp]
\centering
\includegraphics[width=\linewidth,height=0.62\textheight,keepaspectratio]{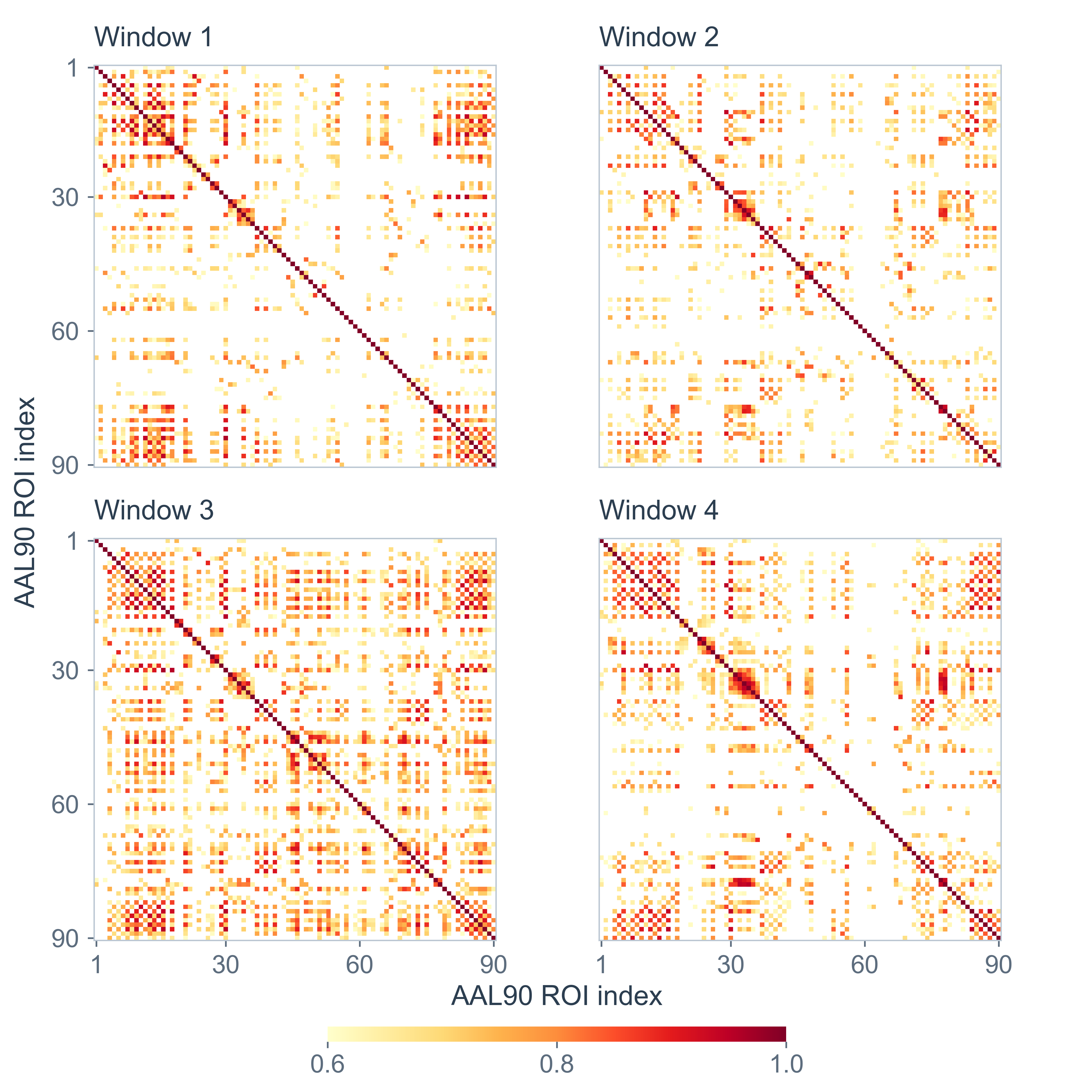}
\caption{Dynamic functional brain networks constructed using ST2GCN. Windows 1--4 show the functional connectivity matrices of the same participant across four consecutive, non-overlapping temporal windows. Rows and columns correspond to the 90 AAL brain regions. Colors represent absolute Pearson correlation coefficients retained at a threshold of 0.6; white cells indicate connections set to zero after thresholding. Diagonal self-connections are retained. All panels share the same ROI ordering and color scale.}
\label{fig:4-1}
\end{figure}

The resulting 4 \(\times\) 90 \(\times\) 90 connectivity sequence and the corresponding 4 \(\times\) 90 \(\times\) 49 windowed ROI signals jointly form the inputs to ST2GCN. The model subsequently organizes node signals and connectivity information using eight sets of functional subnetwork indices that allow overlapping membership. It extracts within-window subnetwork topological features and relates representations of the same brain region across windows. This visualization presents the input networks generated during network construction and connects the specific temporal ranges and regional connectivity distributions with subsequent feature learning. The next section summarizes classification performance for the 27 methods, including ST2GCN, using their respective network inputs to examine the toolkit's support for practical brain network classification tasks.

\subsection{Classification Performance of 27 Methods under Reference Configurations}

Based on the NC--AD binary classification task described in Section 4.1, this section evaluates the classification performance of the 27 brain network analysis methods integrated into the toolkit under reference configurations. The evaluation covers four categories: static association and temporally aggregated topological methods, dynamic temporal methods, complex spatiotemporal and higher-order topological methods, and joint structure--function methods. The complete classification results are presented in Table 4-1. Across the methods, mean ACC ranges from 0.7162 to 0.9510, and mean AUC ranges from 0.7646 to 0.9530. The workflows show different patterns of overall accuracy, discrimination, and class-specific identification, providing a reference for selecting analytical methods according to the research task.

\begin{table}[!htbp]
\centering
\begingroup
\fontsize{8}{10}\selectfont
\setlength{\tabcolsep}{2pt}
\renewcommand{\arraystretch}{1.15}
\caption{Classification performance of 27 methods on the ADNI dataset (mean $\pm$ standard deviation).}\label{tab:classification}
\begin{tabular}{@{}>{\raggedright\arraybackslash}p{1.10in}>{\raggedright\arraybackslash}p{0.95in}*{5}{c}@{}}
\toprule
\textbf{Category} & \textbf{Method} & \textbf{ACC} & \textbf{AUC} & \textbf{F1} & \textbf{SEN} & \textbf{SPE} \\
\midrule
\multirow[t]{13}{=}{Static association and temporally aggregated topological methods} & SVM & $0.7162\pm0.1268$ & $0.8571\pm0.1044$ & $0.6625\pm0.1805$ & $0.6143\pm0.2124$ & $0.8107\pm0.1580$ \\*
 & LDA & $0.7857\pm0.0987$ & $0.8495\pm0.0994$ & $0.7552\pm0.1449$ & $0.7286\pm0.1964$ & $0.8393\pm0.1365$ \\*
 & Logistic Regression & $0.7924\pm0.1077$ & $0.8582\pm0.0980$ & $0.7755\pm0.1221$ & $0.7571\pm0.1571$ & $0.8232\pm0.1545$ \\*
 & CCA--LDA & $0.7924\pm0.1237$ & $0.7902\pm0.1242$ & $0.7791\pm0.1351$ & $0.7714\pm0.1714$ & $0.8089\pm0.1698$ \\*
 & Random Forest & $0.7581\pm0.1233$ & $0.8616\pm0.1042$ & $0.7112\pm0.1681$ & $0.6429\pm0.1835$ & $0.8661\pm0.1327$ \\*
 & XGBoost & $0.7581\pm0.1055$ & $0.8464\pm0.1047$ & $0.7035\pm0.1687$ & $0.6429\pm0.2045$ & $0.8643\pm0.1163$ \\*
 & MLP & $0.7781\pm0.1287$ & $0.8531\pm0.0974$ & $0.7685\pm0.1383$ & $0.7714\pm0.1714$ & $0.7804\pm0.1828$ \\*
 & SimSiam & $0.7995\pm0.1178$ & $0.8794\pm0.1542$ & $0.7587\pm0.1577$ & $0.7000\pm0.2162$ & $0.8911\pm0.1499$ \\*
 & BrainNetCNN & $0.8686\pm0.0771$ & $0.8839\pm0.0858$ & $0.8498\pm0.0976$ & $0.8143\pm0.1571$ & $0.9161\pm0.0687$ \\*
 & GCN & $0.7776\pm0.0773$ & $0.8347\pm0.1012$ & $0.7203\pm0.1289$ & $0.6286\pm0.1714$ & $0.9179\pm0.0674$ \\*
 & GIN & $0.7624\pm0.0998$ & $0.8381\pm0.1192$ & $0.7594\pm0.0814$ & $0.7571\pm0.1436$ & $0.7661\pm0.2483$ \\*
 & GraphSAGE & $0.8062\pm0.0979$ & $0.8219\pm0.1414$ & $0.7450\pm0.1624$ & $0.6571\pm0.2231$ & $0.9464\pm0.0659$ \\*
 & BrainGNN & $0.7557\pm0.0769$ & $0.8158\pm0.0939$ & $0.7401\pm0.0910$ & $0.7286\pm0.1491$ & $0.7768\pm0.1621$ \\
\midrule
\multirow[t]{6}{=}{Dynamic temporal methods} & CNN & $0.8129\pm0.0711$ & $0.8151\pm0.1208$ & $0.7530\pm0.1357$ & $0.6429\pm0.1835$ & $0.9714\pm0.0571$ \\*
 & BiLSTM & $0.8057\pm0.0832$ & $0.7883\pm0.1142$ & $0.7339\pm0.1453$ & $0.6000\pm0.1784$ & $1.0000\pm0.0000$ \\*
 & Transformer & $0.7500\pm0.1218$ & $0.7834\pm0.1325$ & $0.6774\pm0.1719$ & $0.5714\pm0.1917$ & $0.9196\pm0.1075$ \\*
 & GAT & $0.7343\pm0.1039$ & $0.8151\pm0.1106$ & $0.7111\pm0.1253$ & $0.7000\pm0.2065$ & $0.7571\pm0.2788$ \\*
 & STGCN & $0.7705\pm0.0955$ & $0.8510\pm0.1200$ & $0.7352\pm0.1206$ & $0.6857\pm0.1784$ & $0.8500\pm0.1072$ \\*
 & STAGIN & $0.7714\pm0.1568$ & $0.8426\pm0.1205$ & $0.7526\pm0.1779$ & $0.7429\pm0.2100$ & $0.7946\pm0.2314$ \\
\midrule
\multirow[t]{5}{=}{Complex spatiotemporal and higher-order topological methods} & ST2GCN & $0.9367\pm0.0499$ & $0.9530\pm0.0478$ & $0.9303\pm0.0508$ & $0.8838\pm0.0845$ & $0.9833\pm0.0500$ \\*
 & CD-DSTCN & $0.8605\pm0.1087$ & $0.9060\pm0.0896$ & $0.8231\pm0.1440$ & $0.7579\pm0.1979$ & $0.9689\pm0.0653$ \\*
 & LR-STIGCN & $0.9229\pm0.0584$ & $0.9428\pm0.0460$ & $0.9228\pm0.0584$ & $0.9274\pm0.0959$ & $0.9278\pm0.1293$ \\*
 & STHAN & $0.9510\pm0.0322$ & $0.9277\pm0.0981$ & $0.9387\pm0.0558$ & $0.9006\pm0.1000$ & $0.9857\pm0.0429$ \\*
 & NeuroH-TGL & $0.7500\pm0.1118$ & $0.7646\pm0.1568$ & $0.5861\pm0.2620$ & $0.4600\pm0.2501$ & $1.0000\pm0.0000$ \\
\midrule
\multirow[t]{3}{=}{Joint structure--function methods} & OT-MCSTGCN & $0.9175\pm0.0297$ & $0.8906\pm0.0958$ & $0.9019\pm0.0538$ & $0.8534\pm0.1067$ & $0.9571\pm0.0915$ \\*
 & MSTGAC & $0.8257\pm0.0983$ & $0.8331\pm0.1482$ & $0.8052\pm0.1130$ & $0.8016\pm0.1921$ & $0.8256\pm0.2937$ \\*
 & ADRNet & $0.8105\pm0.1014$ & $0.8827\pm0.1116$ & $0.7934\pm0.1038$ & $0.7429\pm0.1245$ & $0.8750\pm0.1457$ \\
\bottomrule
\end{tabular}
\par\smallskip
\noindent\parbox{\linewidth}{Note: ACC, accuracy; AUC, area under the receiver operating characteristic curve; F1, F1 score; SEN, sensitivity; SPE, specificity.}
\endgroup
\end{table}

Static association and temporally aggregated topological methods include conventional statistical learning, machine learning, and aggregated graph representation workflows. Among the conventional statistical learning and machine learning methods, mean ACC ranges from 0.7162 to 0.7924. Logistic Regression achieves an ACC of 0.7924\(\pm\)0.1077 and an AUC of 0.8582\(\pm\)0.0980, providing a connectivity-based classification reference for the current task. Among methods using temporally aggregated graph topology, BrainNetCNN achieves an ACC of 0.8686\(\pm\)0.0771, an AUC of 0.8839\(\pm\)0.0858, and an F1 score of 0.8498\(\pm\)0.0976, with the highest mean for each of these metrics in this category. GraphSAGE achieves an ACC of 0.8062\(\pm\)0.0979. SimSiam achieves an ACC of 0.7995\(\pm\)0.1178 and an AUC of 0.8794\(\pm\)0.1542, with its mean AUC ranking second to BrainNetCNN in this category. These results illustrate multiple classification approaches, from connectivity edge features to aggregated graph topology representations, and show how different feature encoding strategies perform in the current task.

Dynamic temporal methods primarily extract temporal features from windowed node signals or dynamic functional graph sequences. Their mean ACC values range from 0.7343 to 0.8129, and their mean AUC values range from 0.7834 to 0.8510. CNN achieves the highest mean ACC in this category, at 0.8129\(\pm\)0.0711. STGCN achieves the highest mean AUC, at 0.8510\(\pm\)0.1200, together with an ACC of 0.7705\(\pm\)0.0955. STAGIN achieves an ACC of 0.7714\(\pm\)0.1568 and an AUC of 0.8426\(\pm\)0.1205. These methods show different strengths in overall classification accuracy and discrimination based on continuous prediction scores, offering several ways to use node activity and evolving connectivity for classification. Analyzing results across methods helps characterize the dependence of dynamic functional connectivity findings on methodological choices and supports the interpretation of workflow outputs in relation to data characteristics and research objectives (Lurie et al., 2020).

Several complex spatiotemporal and higher-order topological methods achieve relatively high classification metrics in this evaluation. STHAN achieves an ACC of 0.9510\(\pm\)0.0322 and an F1 score of 0.9387\(\pm\)0.0558, with the highest means for both metrics across all methods. Its AUC is 0.9277\(\pm\)0.0981. ST2GCN achieves the highest mean AUC across all methods, at 0.9530\(\pm\)0.0478, together with an ACC of 0.9367\(\pm\)0.0499 and an F1 score of 0.9303\(\pm\)0.0508. Its mean ACC and F1 score both rank second to STHAN. LR-STIGCN achieves an AUC of 0.9428\(\pm\)0.0460, with the second-highest mean across all methods, together with an ACC of 0.9229\(\pm\)0.0584 and an F1 score of 0.9228\(\pm\)0.0584. These results illustrate the performance of the complex spatiotemporal and higher-order topological workflows integrated into the toolkit in the current classification task.

Joint structure--function methods use paired fMRI and DTI data for classification, demonstrating the toolkit's support for multimodal brain network analysis. In this category, OT-MCSTGCN achieves an ACC of 0.9175\(\pm\)0.0297 and an AUC of 0.8906\(\pm\)0.0958, with the highest mean ACC in this category. MSTGAC achieves an ACC of 0.8257\(\pm\)0.0983 and an AUC of 0.8331\(\pm\)0.1482, while ADRNet achieves an ACC of 0.8105\(\pm\)0.1014 and an AUC of 0.8827\(\pm\)0.1116. MSTGAC has the second-highest mean ACC in this category, whereas OT-MCSTGCN has the highest mean AUC in this category, illustrating the different performance profiles of joint structure--function workflows across evaluation metrics. These results provide a basis for comparing methodological options in studies with paired structural and functional imaging data.

These results should also be interpreted in light of the sample size (Varoquaux, 2018). With 10-fold cross-validation in 143 participants, each test fold contains only 14--15 participants, with class composition depending on the fold-generation strategy used by each method. A change in the classification outcome for a single participant can therefore change ACC in that fold by approximately 6.7--7.1 percentage points. Sensitivity, specificity, and the F1 score may also vary substantially because of the small number of participants in each test fold. These metrics provide a quantitative reference for the predictive performance of different workflows. The next section uses ST2GCN to further examine candidate brain regions and connectivity patterns, connecting classification evaluation with the toolkit's brain biomarker discovery module.

\subsection{Analysis of Candidate Brain Connections and Regions Based on ST2GCN}

To examine the toolkit's support for brain biomarker discovery, this section uses ST2GCN as an example. Candidate brain connections and regions associated with the AD classification task are identified by combining discriminative functional connectivity information estimated using ElasticNet (Zou and Hastie, 2005) with subnetwork attention weights. Importance scores obtained from the ten cross-validation models across all 143 participants were averaged, and the Top-20 connections and regions were selected separately for candidate analysis. These features were anatomically localized using the AAL90 atlas, and their spatial distributions and potential disease associations were examined in the context of existing AD research.

At the connection level, Figure 4-2 shows the spatial distribution and relative importance of the candidate connections. The Top-20 connections involved 25 distinct brain regions. Twelve connections were interhemispheric, accounting for 60\% of the candidate set, while six were within the left hemisphere and two were within the right hemisphere. These connections included frontoparietal, frontotemporal, and intraparietal links. For example, the connections between the left postcentral gyrus and left inferior parietal lobule, between the right middle frontal gyrus and the temporal pole of the left middle temporal gyrus, and between the triangular part of the left inferior frontal gyrus and the left superior parietal lobule ranked third, fourth, and fifth, with importance scores of 0.8643, 0.7622, and 0.7526, respectively. The frontoparietal connections provide a basis for examining the candidate features in relation to attention. Using resting-state functional MRI and longitudinal cognitive assessments from 142 participants, Pahl et al. (2024) found that lower functional connectivity in the left frontoparietal network was associated with poorer performance on attention tasks and faster subsequent decline. Their findings provide background for interpreting the potential cognitive associations of the frontoparietal connections identified here.

\begin{figure}[!htbp]
\centering
\includegraphics[width=\linewidth,height=0.72\textheight,keepaspectratio]{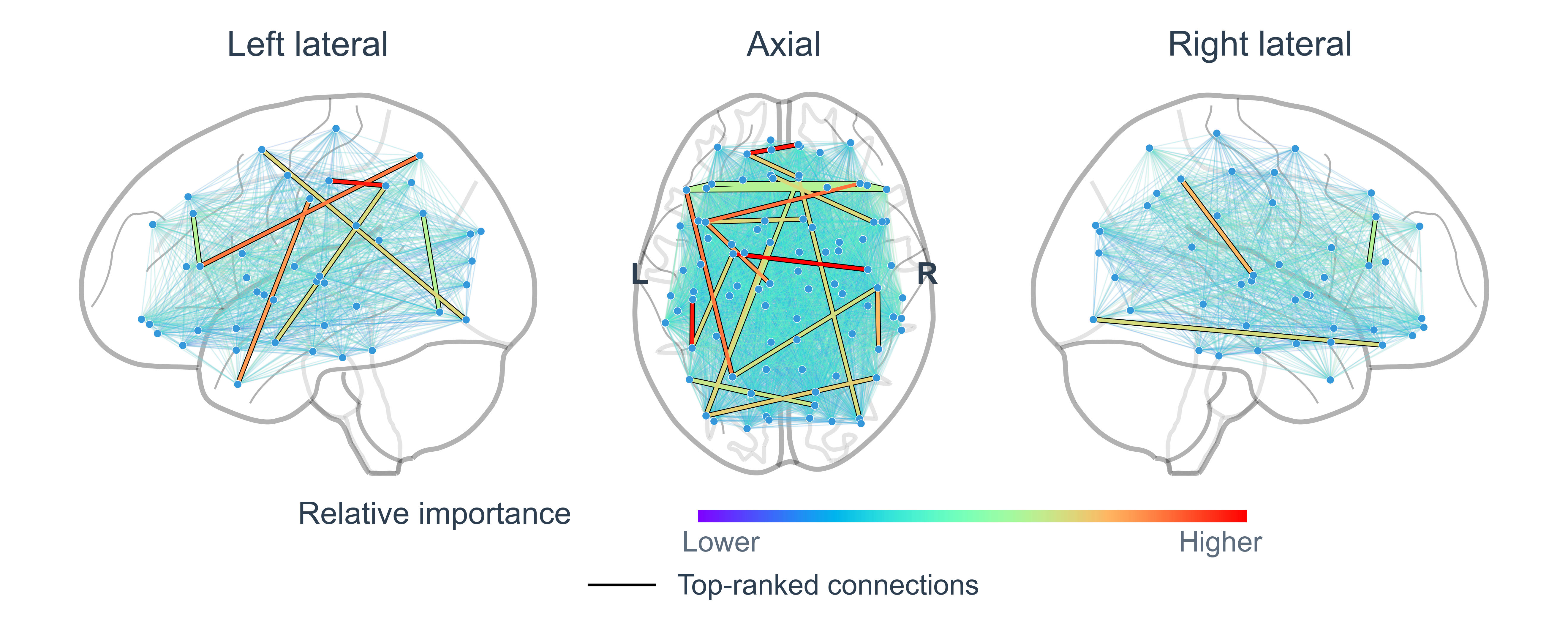}
\caption{Spatial distribution of ST2GCN-derived connection importance. Connections are shown in left lateral, axial, and right lateral views. Rainbow colors indicate relative importance. Black outlines highlight the Top-20 connections; other connections have light-gray outlines. Blue nodes represent AAL regional centroids.}
\label{fig:4-2}
\end{figure}

Connections involving the amygdala, temporal pole, and orbitofrontal regions were also present in the candidate set. The connection between the right precentral gyrus and left amygdala had the highest importance score, 0.8945. The connection between the orbital part of the left superior frontal gyrus and the medial orbital part of the right superior frontal gyrus ranked second, with a score of 0.8692. In addition, connections between the left amygdala and left inferior parietal lobule, between the left midcingulate region and the temporal pole of the left middle temporal gyrus, and between the left gyrus rectus and the temporal pole of the right middle temporal gyrus were among the Top-20. In 261 participants spanning normal aging and the AD continuum, Chauveau et al. (2025) found that increased functional connectivity in the anterior-temporal network was associated with amyloid burden, reduced glucose metabolism, hippocampal atrophy, and cognitive impairment, as well as faster progression to AD dementia among participants with MCI. The anterior-temporal system examined in that study involves the amygdala, temporal pole, and orbitofrontal regions, providing a network-level disease context for the candidate connections identified here. The correspondence primarily concerns the regions and functional systems involved; the specific region pairs remain candidate features selected in the present analysis.

At the regional level, Figure 4-3 shows the spatial distribution and relative importance of the candidate regions. The Top-20 regions comprised ten bilateral pairs of corresponding regions, with ten regions in each hemisphere. According to the current atlas annotations, the frontal, parietal, temporal, and cingulate groups contained eight, six, four, and two candidate regions, respectively. Frontal and parietal regions together accounted for 70\% of the candidate set. The triangular part of the left inferior frontal gyrus, right middle frontal gyrus, left middle frontal gyrus, and triangular part of the right inferior frontal gyrus ranked first through fourth, with importance scores of 0.8684, 0.8681, 0.8647, and 0.8640, respectively. The bilateral inferior parietal lobules and anterior cingulate regions occupied ranks five through eight. This distribution indicates that frontal and parietal regions were prominent in the regional importance ranking obtained here. Using memory-encoding task fMRI from 490 participants, Vockert et al. (2024) identified a brain activity pattern associated with cognitive reserve. This pattern included the bilateral triangular, opercular, and orbital parts of the inferior frontal gyrus, as well as inferior parietal cortex around the angular gyrus. Expression of this pattern was associated with less AD pathology-related cognitive impairment and slower subsequent cognitive decline, providing an interpretive context involving memory processing and cognitive reserve for candidate regions such as the inferior frontal gyrus identified here.

The parietal candidates also included the bilateral supramarginal and angular gyri. The left and right supramarginal gyri ranked eleventh and thirteenth, respectively, while the left and right angular gyri ranked seventeenth and eighteenth. In the connection results, the angular gyrus also formed candidate links with the inferior occipital gyrus, cortex surrounding the calcarine sulcus, and lingual gyrus, illustrating connections between parietal and occipital regions. Guo et al. (2026) reported differences in regional homogeneity involving the angular and supramarginal gyri in relation to AD pathology. This anatomical correspondence provides literature support for the potential AD relevance of the parietal candidates identified in the present analysis.

\begin{figure}[!htbp]
\centering
\includegraphics[width=\linewidth,height=0.72\textheight,keepaspectratio]{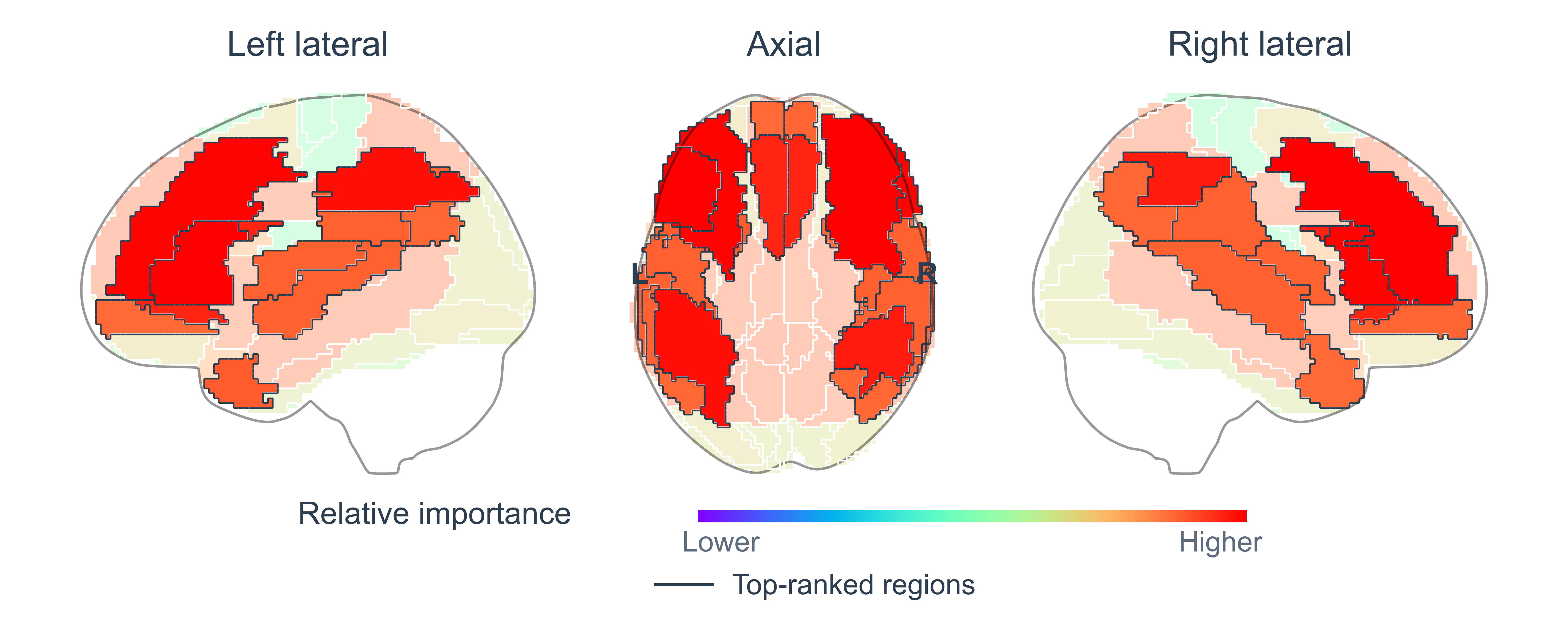}
\caption{Anatomical distribution of ST2GCN-derived regional importance. AAL parcels are shown in left lateral, axial, and right lateral projections. Rainbow colors indicate relative importance. Top-ranked regions have dark outlines and full-opacity colors; other regions appear paler. Where parcels overlap in projection, the highest-scoring parcel determines the displayed color.}
\label{fig:4-3}
\end{figure}

The analyses above present candidate features generated by ST2GCN in the current cohort at both the connection and regional levels, with anatomical localization and recent AD research providing spatial and disease context. Importance scores reflect relative rankings within the current analysis workflow and cannot be directly interpreted as increased or decreased connectivity in the AD group relative to the NC group, or as evidence of established disease biomarkers. By linking importance estimation, candidate selection, spatial visualization, and analysis of disease associations, this example demonstrates the toolkit's support for candidate brain biomarker discovery and interpretation, providing practical evidence for the functional validation of the brain biomarker discovery module.

\section{Discussion}

This study evaluated the classification performance of 27 methods under reference configurations using 10-fold cross-validation in 143 participants from ADNI for NC--AD classification. The evaluation provides a reference for understanding how different analytical workflows perform under specific data conditions. Using ST2GCN as an example, brain network construction and visualization showed connectivity patterns across time windows. Analyses of candidate brain connections and regions further illustrated discriminative features associated with the current classification task and their spatial distributions. Anatomical localization and consideration of relevant studies provide a basis for discussing the potential disease associations of these candidate features and generating leads for further research. These results demonstrate the toolkit's support for practical neuroimaging analysis at the levels of network input, classification performance, and candidate features. Researchers can configure appropriate workflows for their own data and research objectives, provided that the input requirements of the selected method are met, and evaluate classification performance and candidate features in the target dataset.

The LLM reporting module based on structured evidence further supports an integrated understanding of brain network analysis results. Depending on the available data modalities, the module organizes individual- or group-level results on functional connectivity, structural connectivity, and structure--function coupling into structured evidence. The LLM then synthesizes global network properties, regional features, and cross-modal associations to generate supporting analytical reports. This functionality helps connect results across different scales, supporting the joint examination of global network organization and local regional characteristics. The structured evidence retains the corresponding metric values, statistical tests, and result sources. Users can inspect the input summary before report generation and subsequently verify the report against the original analysis results, supporting the traceability of the evidence underlying assisted interpretations and the verification of results.

This study has several limitations. Current application validation focuses on NC--AD classification using ADNI data. The reported performance corresponds to the data splits and reference configurations used in this study, and applicability under other data conditions and research tasks requires further investigation (Scheinost et al., 2019). Given the limited sample size, each test fold in 10-fold cross-validation contains relatively few participants. Consequently, estimates of classification metrics can be affected by changes in predictions for a small number of participants (Varoquaux, 2018). The brain regions and connections identified through importance analysis are candidate features obtained within the current workflow, and their disease associations require support from independent data and further research. In addition, the LLM reporting module currently demonstrates its functionality and usage workflow, while independent evaluation of report quality remains a direction for future work. Future work could extend application validation to additional datasets and tasks, examine the reproducibility of candidate features, and combine expert evaluation with verification against source data to assess the consistency between reports and statistical evidence, the extent of information coverage, and the appropriateness of assisted interpretations (Gao et al., 2023). These efforts could guide continued improvements to the toolkit's analysis and reporting functions.

\section{Conclusions}

BrainNet Studio provides a user-friendly and extensible toolkit for static and dynamic brain network analysis. It establishes a unified workflow spanning network construction and visualization, feature extraction, predictive modeling, and the identification of candidate brain regions and connections. Through its graphical interface, researchers can import brain signal data, configure analytical tasks, train models, and inspect and export results without manually assembling multiple processing pipelines. The LLM-based reporting module serves as a complementary workflow that organizes structured evidence to assist the interpretation of individual- and group-level findings. Collectively, these capabilities position BrainNet Studio as a configurable platform for intelligent brain network construction, analysis, and interpretation in cognitive neuroscience, exploratory research on brain disorders, and brain--computer interface studies. 

\section*{Data availability}

The fMRI and DTI data used in this study were obtained from the Alzheimer's Disease Neuroimaging Initiative (ADNI) database. These data are available to approved researchers through the LONI Image and Data Archive (IDA), subject to the ADNI Data Use Agreement and applicable data sharing policies. Instructions for requesting access are provided at \url{https://adni.loni.usc.edu/data-samples/adni-data/}.

\section*{Code availability}

The source code for the toolkit is publicly available at \url{https://github.com/xbrainnet/Brainnet-Studio}.

\FloatBarrier
\setlength{\bibsep}{2pt}


\begin{thebibliography}{50}

\bibitem[Bassett and Sporns(2017)]{bassett2017}
Bassett, D.S., Sporns, O., 2017. Network neuroscience. Nat. Neurosci. 20, 353--364. DOI: 10.1038/nn.4502.

\bibitem[Benjamini and Hochberg(1995)]{benjamini1995}
Benjamini, Y., Hochberg, Y., 1995. Controlling the False Discovery Rate: A Practical and Powerful Approach to Multiple Testing. J. R. Stat. Soc. Series B Methodol. 57, 289--300. DOI: 10.1111/j.2517-6161.1995.tb02031.x.

\bibitem[Chauveau et al.(2025)]{chauveau2025}
Chauveau, L., Landeau, B., Dautricourt, S., Turpin, A.-L., Delarue, M., H{\'{e}}bert, O., de La Sayette, V., Ch{\'{e}}telat, G., de Flores, R., 2025. Anterior-temporal network hyperconnectivity is key to Alzheimer's disease: from ageing to dementia. Brain 148, 2008--2022. DOI: 10.1093/brain/awaf008.

\bibitem[Cui et al.(2026)]{cui2026}
Cui, J., Ye, W., Li, S., Wen, J., Zhu, Q., 2026. Adjacent-Aware Modality Recovery Based on Incomplete Multi-Modal Brain Disease Diagnosis. IEEE Trans. Med. Imaging 45, 2373--2384. DOI: 10.1109/TMI.2026.3654000.

\bibitem[Cui et al.(2013)]{cui2013}
Cui, Z., Zhong, S., Xu, P., He, Y., Gong, G., 2013. PANDA: a pipeline toolbox for analyzing brain diffusion images. Front. Hum. Neurosci. 7, 42. DOI: 10.3389/fnhum.2013.00042.

\bibitem[Fotiadis et al.(2024)]{fotiadis2024}
Fotiadis, P., Parkes, L., Davis, K.A., Satterthwaite, T.D., Shinohara, R.T., Bassett, D.S., 2024. Structure--function coupling in macroscale human brain networks. Nat. Rev. Neurosci. 25, 688--704. DOI: 10.1038/s41583-024-00846-6.

\bibitem[Gadgil et al.(2020)]{gadgil2020}
Gadgil, S., Zhao, Q., Pfefferbaum, A., Sullivan, E.V., Adeli, E., Pohl, K.M., 2020. Spatio-Temporal Graph Convolution for Resting-State fMRI Analysis. In: Medical Image Computing and Computer Assisted Intervention -- MICCAI 2020, Lecture Notes in Computer Science, vol. 12267. Springer, Cham, pp. 528--538. DOI: 10.1007/978-3-030-59728-3\_52. arXiv:2003.10613.

\bibitem[Gao et al.(2023)]{gao2023}
Gao, T., Yen, H., Yu, J., Chen, D., 2023. Enabling Large Language Models to Generate Text with Citations. In: Bouamor, H., Pino, J., Bali, K. (Eds.), Proceedings of the 2023 Conference on Empirical Methods in Natural Language Processing. Association for Computational Linguistics, Singapore, pp. 6465--6488. DOI: 10.18653/v1/2023.emnlp-main.398. arXiv:2305.14627.

\bibitem[Guo et al.(2026)]{guo2026}
Guo, W., Shao, H., Zhang, Y., Liu, L., Zheng, H., Liang, D., Liu, J., Zhang, L., Hu, Z., 2026. Association of functional brain alterations with \(\beta\)-amyloid, tau, and cognitive decline in Alzheimer's disease. Alzheimers Res. Ther. 18, 74. DOI: 10.1186/s13195-026-01991-z.

\bibitem[Kan et al.(2022)]{kan2022}
Kan, X., Dai, W., Cui, H., Zhang, Z., Guo, Y., Yang, C., 2022. Brain Network Transformer. In: Advances in Neural Information Processing Systems 35, pp. 25586--25599. DOI: 10.52202/068431-1855. arXiv:2210.06681.

\bibitem[Kawahara et al.(2017)]{kawahara2017}
Kawahara, J., Brown, C.J., Miller, S.P., Booth, B.G., Chau, V., Grunau, R.E., Zwicker, J.G., Hamarneh, G., 2017. BrainNetCNN: Convolutional neural networks for brain networks; towards predicting neurodevelopment. NeuroImage 146, 1038--1049. DOI: 10.1016/j.neuroimage.2016.09.046.

\bibitem[Kim and Ye(2020)]{kim2020}
Kim, B.-H., Ye, J.C., 2020. Understanding Graph Isomorphism Network for rs-fMRI Functional Connectivity Analysis. Front. Neurosci. 14, 630. DOI: 10.3389/fnins.2020.00630. arXiv:2001.03690.

\bibitem[Kim et al.(2021)]{kim2021}
Kim, B.-H., Ye, J.C., Kim, J.-J., 2021. Learning Dynamic Graph Representation of Brain Connectome with Spatio-Temporal Attention. In: Advances in Neural Information Processing Systems 34, pp. 4314--4327. arXiv:2105.13495.

\bibitem[Kim et al.(2016)]{kim2016}
Kim, J., Calhoun, V.D., Shim, E., Lee, J.-H., 2016. Deep neural network with weight sparsity control and pre-training extracts hierarchical features and enhances classification performance: Evidence from whole-brain resting-state functional connectivity patterns of schizophrenia. NeuroImage 124, 127--146. DOI: 10.1016/j.neuroimage.2015.05.018.

\bibitem[Ktena et al.(2017)]{ktena2017}
Ktena, S.I., Parisot, S., Ferrante, E., Rajchl, M., Lee, M., Glocker, B., Rueckert, D., 2017. Distance Metric Learning Using Graph Convolutional Networks: Application to Functional Brain Networks. In: Medical Image Computing and Computer Assisted Intervention -- MICCAI 2017, Lecture Notes in Computer Science, vol. 10433. Springer, Cham, pp. 469--477. DOI: 10.1007/978-3-319-66182-7\_54. arXiv:1703.02161.

\bibitem[Li et al.(2026)]{li2026}
Li, C., Gong, P., Li, S., Tian, C., Yu, Y., Wang, R., Zhang, D., Zhu, Q., 2026. Spatio-Temporal Hypergraph Attention Networks for Brain Disease Analysis. IEEE Trans. Image Process. 35, 2727--2739. DOI: 10.1109/TIP.2026.3671657.

\bibitem[Li et al.(2025a)]{li2025a}
Li, C., Ma, K., Li, S., Meng, X., Wang, R., Zhang, D., Zhu, Q., 2025a. Multi-channel spatio-temporal graph attention contrastive network for brain disease diagnosis. NeuroImage 307, 121013. DOI: 10.1016/j.neuroimage.2025.121013.

\bibitem[Li et al.(2025b)]{li2025b}
Li, S., Zhu, Q., Guan, D., Shen, B., Zhang, L., Ji, Y., Qi, S., Zhang, D., 2025b. Long-Interval Spatio-Temporal Graph Convolution for Brain Disease Diagnosis. IEEE Trans. Instrum. Meas. 74, 1--11, Art. no. 4004511. DOI: 10.1109/TIM.2025.3551032.

\bibitem[Li et al.(2025c)]{li2025c}
Li, S., Zhu, Q., Tian, C., Shao, W., Zhang, D., 2025c. Interpretable Dynamic Brain Network Analysis With Functional and Structural Priors. IEEE Trans. Med. Imaging 44, 4878--4889. DOI: 10.1109/TMI.2025.3584231.

\bibitem[Li et al.(2025d)]{li2025d}
Li, S., Zhu, Q., Tian, C., Zhang, X., Shao, W., Wen, J., Zhang, D., 2025d. NeuroH-TGL: Neuro-Heterogeneity Guided Temporal Graph Learning Strategy for Brain Disease Diagnosis. In: Advances in Neural Information Processing Systems 38, pp. 121069--121091. DOI: 10.52202/085713-3647.

\bibitem[Li et al.(2021)]{li2021}
Li, X., Zhou, Y., Dvornek, N., Zhang, M., Gao, S., Zhuang, J., Scheinost, D., Staib, L.H., Ventola, P., Duncan, J.S., 2021. BrainGNN: Interpretable Brain Graph Neural Network for fMRI Analysis. Med. Image Anal. 74, 102233. DOI: 10.1016/j.media.2021.102233.

\bibitem[Lin et al.(2020)]{lin2020}
Lin, W., Lv, D., Han, Z., Dong, J., Yang, L., 2020. Major depressive disorder identification by referenced multiset canonical correlation analysis with clinical scores. Med. Image Anal. 60, 101600. DOI: 10.1016/j.media.2019.101600.

\bibitem[Lundberg and Lee(2017)]{lundberg2017}
Lundberg, S.M., Lee, S.-I., 2017. A unified approach to interpreting model predictions. In: Advances in Neural Information Processing Systems 30, pp. 4765--4774. arXiv:1705.07874.

\bibitem[Lurie et al.(2020)]{lurie2020}
Lurie, D.J., Kessler, D., Bassett, D.S., Betzel, R.F., Breakspear, M., Keilholz, S., Kucyi, A., Li{\'{e}}geois, R., Lindquist, M.A., McIntosh, A.R., Poldrack, R.A., Shine, J.M., Thompson, W.H., Bielczyk, N.Z., Douw, L., Kraft, D., Miller, R.L., Muthuraman, M., Pasquini, L., Razi, A., Vidaurre, D., Xie, H., Calhoun, V.D., 2020. Questions and controversies in the study of time-varying functional connectivity in resting fMRI. Netw. Neurosci. 4, 30--69. DOI: 10.1162/netn\_a\_00116.

\bibitem[McCaw et al.(2020)]{mccaw2020}
McCaw, Z.R., Lane, J.M., Saxena, R., Redline, S., Lin, X., 2020. Operating characteristics of the rank-based inverse normal transformation for quantitative trait analysis in genome-wide association studies. Biometrics 76, 1262--1272. DOI: 10.1111/biom.13214.

\bibitem[Meszl{\'{e}}nyi et al.(2017)]{meszlenyi2017}
Meszl{\'{e}}nyi, R.J., Buza, K., Vidny{\'{a}}nszky, Z., 2017. Resting State fMRI Functional Connectivity-Based Classification Using a Convolutional Neural Network Architecture. Front. Neuroinform. 11, 61. DOI: 10.3389/fninf.2017.00061.

\bibitem[Meunier et al.(2020)]{meunier2020}
Meunier, D., Pascarella, A., Altukhov, D., Jas, M., Combrisson, E., Lajnef, T., Bertrand-Dubois, D., Hadid, V., Alamian, G., Alves, J., Barlaam, F., Saive, A.-L., Dehgan, A., Jerbi, K., 2020. NeuroPycon: An open-source python toolbox for fast multi-modal and reproducible brain connectivity pipelines. NeuroImage 219, 117020. DOI: 10.1016/j.neuroimage.2020.117020.

\bibitem[Mijalkov et al.(2017)]{mijalkov2017}
Mijalkov, M., Kakaei, E., Pereira, J.B., Westman, E., Volpe, G., for the Alzheimer's Disease Neuroimaging Initiative, 2017. BRAPH: A graph theory software for the analysis of brain connectivity. PLoS One 12, e0178798. DOI: 10.1371/journal.pone.0178798.

\bibitem[Pahl et al.(2024)]{pahl2024}
Pahl, J., Prokopiou, P.C., Bueichek{\'{u}}, E., Schultz, A.P., Papp, K.V., Farrell, M.E., Rentz, D.M., Sperling, R.A., Johnson, K.A., Jacobs, H.I.L., 2024. Locus coeruleus integrity and left frontoparietal connectivity provide resilience against attentional decline in preclinical Alzheimer's disease. Alzheimers Res. Ther. 16, 119. DOI: 10.1186/s13195-024-01485-w.

\bibitem[Rubinov and Sporns(2010)]{rubinov2010}
Rubinov, M., Sporns, O., 2010. Complex network measures of brain connectivity: uses and interpretations. NeuroImage 52, 1059--1069. DOI: 10.1016/j.neuroimage.2009.10.003.

\bibitem[Scheinost et al.(2019)]{scheinost2019}
Scheinost, D., Noble, S., Horien, C., Greene, A.S., Lake, E.M.R., Salehi, M., Gao, S., Shen, X., O'Connor, D., Barron, D.S., Yip, S.W., Rosenberg, M.D., Constable, R.T., 2019. Ten simple rules for predictive modeling of individual differences in neuroimaging. NeuroImage 193, 35--45. DOI: 10.1016/j.neuroimage.2019.02.057.

\bibitem[Shevchenko et al.(2025)]{shevchenko2025}
Shevchenko, V., Benn, R.A., Scholz, R., Wei, W., Pallavicini, C., Klatzmann, U., Alberti, F., Satterthwaite, T.D., Wassermann, D., Bazin, P.-L., Margulies, D.S., 2025. A comparative machine learning study of schizophrenia biomarkers derived from functional connectivity. Sci. Rep. 15, 2849. DOI: 10.1038/s41598-024-84152-2.

\bibitem[Singhal et al.(2023)]{singhal2023}
Singhal, K., Azizi, S., Tu, T., Mahdavi, S.S., Wei, J., Chung, H.W., Scales, N., Tanwani, A., Cole-Lewis, H., Pfohl, S., Payne, P., Seneviratne, M., Gamble, P., Kelly, C., Babiker, A., Sch{\"{a}}rli, N., Chowdhery, A., Mansfield, P., Demner-Fushman, D., Ag{\"{u}}era y Arcas, B., Webster, D., Corrado, G.S., Matias, Y., Chou, K., Gottweis, J., Tomasev, N., Liu, Y., Rajkomar, A., Barral, J., Semturs, C., Karthikesalingam, A., Natarajan, V., 2023. Large language models encode clinical knowledge. Nature 620, 172--180. DOI: 10.1038/s41586-023-06291-2.

\bibitem[Tzourio-Mazoyer et al.(2002)]{tzouriomazoyer2002}
Tzourio-Mazoyer, N., Landeau, B., Papathanassiou, D., Crivello, F., Etard, O., Delcroix, N., Mazoyer, B., Joliot, M., 2002. Automated anatomical labeling of activations in SPM using a macroscopic anatomical parcellation of the MNI MRI single-subject brain. NeuroImage 15, 273--289. DOI: 10.1006/nimg.2001.0978.

\bibitem[Varoquaux(2018)]{varoquaux2018}
Varoquaux, G., 2018. Cross-validation failure: Small sample sizes lead to large error bars. NeuroImage 180, 68--77. DOI: 10.1016/j.neuroimage.2017.06.061.

\bibitem[Vockert et al.(2024)]{vockert2024}
Vockert, N., Machts, J., Kleineidam, L., Nemali, A., Incesoy, E.I., Bernal, J., Sch{\"{u}}tze, H., Yakupov, R., Peters, O., Gref, D., Schneider, L.S., Preis, L., Priller, J., Spruth, E.J., Altenstein, S., Schneider, A., Fliessbach, K., Wiltfang, J., Rostamzadeh, A., Glanz, W., Teipel, S., Kilimann, I., Goerss, D., Laske, C., Munk, M.H., Spottke, A., Roy, N., Heneka, M.T., Brosseron, F., Wagner, M., Wolfsgruber, S., Dobisch, L., Dechent, P., Hetzer, S., Scheffler, K., Zeidman, P., Stern, Y., Schott, B.H., Jessen, F., D{\"{u}}zel, E., Maass, A., Ziegler, G., the DELCODE study group, 2024. Cognitive reserve against Alzheimer's pathology is linked to brain activity during memory formation. Nat. Commun. 15, 9815. DOI: 10.1038/s41467-024-53360-9.

\bibitem[Waller et al.(2018)]{waller2018}
Waller, L., Brovkin, A., Dorfschmidt, L., Bzdok, D., Walter, H., Kruschwitz, J.D., 2018. GraphVar 2.0: A user-friendly toolbox for machine learning on functional connectivity measures. J. Neurosci. Methods 308, 21--33. DOI: 10.1016/j.jneumeth.2018.07.001.

\bibitem[Wang et al.(2022)]{wang2022}
Wang, D., Wu, Q., Hong, D., 2022. Extracting default mode network based on graph neural network for resting state fMRI study. Front. Neuroimaging 1, 963125. DOI: 10.3389/fnimg.2022.963125.

\bibitem[Wang et al.(2015)]{wang2015}
Wang, J., Wang, X., Xia, M., Liao, X., Evans, A., He, Y., 2015. GRETNA: a graph theoretical network analysis toolbox for imaging connectomics. Front. Hum. Neurosci. 9, 386. DOI: 10.3389/fnhum.2015.00386.

\bibitem[Wang et al.(2025)]{wang2025}
Wang, X., Fang, Y., Wang, Q., Yap, P.-T., Zhu, H., Liu, M., 2025. Self-supervised graph contrastive learning with diffusion augmentation for functional MRI analysis and brain disorder detection. Med. Image Anal. 101, 103403. DOI: 10.1016/j.media.2024.103403.

\bibitem[Weiner et al.(2025)]{weiner2025}
Weiner, M.W., Kanoria, S., Miller, M.J., Aisen, P.S., Beckett, L.A., Conti, C., Diaz, A., Flenniken, D., Green, R.C., Harvey, D.J., Jack, C.R., Jr., Jagust, W., Lee, E.B., Morris, J.C., Nho, K., Nosheny, R., Okonkwo, O.C., Perrin, R.J., Petersen, R.C., Rivera-Mindt, M., Saykin, A.J., Shaw, L.M., Toga, A.W., Tosun, D., Veitch, D.P., for the Alzheimer's Disease Neuroimaging Initiative, 2025. Overview of Alzheimer's Disease Neuroimaging Initiative and future clinical trials. Alzheimers Dement. 21, e14321. DOI: 10.1002/alz.14321.

\bibitem[Whitfield-Gabrieli and Nieto-Castanon(2012)]{whitfieldgabrieli2012}
Whitfield-Gabrieli, S., Nieto-Castanon, A., 2012. Conn: a functional connectivity toolbox for correlated and anticorrelated brain networks. Brain Connect. 2, 125--141. DOI: 10.1089/brain.2012.0073.

\bibitem[Wu et al.(2023)]{wu2023}
Wu, J., Fang, Y., Tan, X., Kang, S., Yue, X., Rao, Y., Huang, H., Liu, M., Qiu, S., Yap, P.-T., 2023. Detecting type 2 diabetes mellitus cognitive impairment using whole-brain functional connectivity. Sci. Rep. 13, 3940. DOI: 10.1038/s41598-023-28163-5.

\bibitem[Yan et al.(2018)]{yan2018}
Yan, W., Zhang, H., Sui, J., Shen, D., 2018. Deep Chronnectome Learning via Full Bidirectional Long Short-Term Memory Networks for MCI Diagnosis. In: Medical Image Computing and Computer Assisted Intervention -- MICCAI 2018, Lecture Notes in Computer Science, vol. 11072. Springer, Cham, pp. 249--257. DOI: 10.1007/978-3-030-00931-1\_29.

\bibitem[Yang et al.(2019)]{yang2019}
Yang, H., Li, X., Wu, Y., Li, S., Lu, S., Duncan, J.S., Gee, J.C., Gu, S., 2019. Interpretable Multimodality Embedding of Cerebral Cortex Using Attention Graph Network for Identifying Bipolar Disorder. In: Medical Image Computing and Computer Assisted Intervention -- MICCAI 2019, Lecture Notes in Computer Science, vol. 11766. Springer, Cham, pp. 799--807. DOI: 10.1007/978-3-030-32248-9\_89.

\bibitem[Yuan et al.(2025)]{yuan2025}
Yuan, N., Guan, D., Li, S., Zhang, L., Zhu, Q., 2025. Enhancing Neurodegenerative Disease Diagnosis Through Confidence-Driven Dynamic Spatio-Temporal Convolutional Network. IEEE Trans. Neural Syst. Rehabil. Eng. 33, 1715--1728. DOI: 10.1109/TNSRE.2025.3564983.

\bibitem[Zalesky et al.(2010)]{zalesky2010}
Zalesky, A., Fornito, A., Bullmore, E.T., 2010. Network-based statistic: Identifying differences in brain networks. NeuroImage 53, 1197--1207. DOI: 10.1016/j.neuroimage.2010.06.041.

\bibitem[Zhou et al.(2020)]{zhou2020}
Zhou, Z., Chen, X., Zhang, Y., Hu, D., Qiao, L., Yu, R., Yap, P.-T., Pan, G., Zhang, H., Shen, D., 2020. A toolbox for brain network construction and classification (BrainNetClass). Hum. Brain Mapp. 41, 2808--2826. DOI: 10.1002/hbm.24979.

\bibitem[Zhu et al.(2024)]{zhu2024}
Zhu, Q., Li, S., Meng, X., Xu, Q., Zhang, Z., Shao, W., Zhang, D., 2024. Spatio-Temporal Graph Hubness Propagation Model for Dynamic Brain Network Classification. IEEE Trans. Med. Imaging 43, 2381--2394. DOI: 10.1109/TMI.2024.3363014.

\bibitem[Zou and Hastie(2005)]{zou2005}
Zou, H., Hastie, T., 2005. Regularization and Variable Selection Via the Elastic Net. J. R. Stat. Soc. Series B Stat. Methodol. 67, 301--320. DOI: 10.1111/j.1467-9868.2005.00503.x.

\end{thebibliography}
\end{document}